\documentclass[10pt,journal,compsoc]{IEEEtran}

\usepackage{ragged2e}

\usepackage{graphicx}
\usepackage{epstopdf}
\usepackage{multirow}
\usepackage{indentfirst}
\usepackage{subfigure}
\usepackage{algorithm}
\usepackage{algpseudocode}
\usepackage{amsmath}
\usepackage{amssymb}
\usepackage{float}
\usepackage{caption2}

\ifCLASSOPTIONcompsoc

  \usepackage[nocompress]{cite}
\else

  \usepackage{cite}
\fi

\ifCLASSINFOpdf

\else

\fi

\begin{document}

\title{An Improved multi-objective genetic algorithm based on orthogonal design and adaptive clustering pruning strategy}

\author{Xinwu~Yang,
		 Guizeng~You,
        Chong~Zhao,
	   Mengfei~Dou
        and~Xinian~Guo
\IEEEcompsocitemizethanks{\IEEEcompsocthanksitem Xinwu. Yang is with the Department of Faculty of Information Technology, Beijing Municipal Key Lab of Multimedia and Intelligent Software Technology. Beijing Key Laboratory on Integration and Analysis of Large Scale Stream Data. Beijing University of Technology, No.100, Pingleyuan, Chaoyang,District, Beijing 100124, China.\protect\\

E-mail: yang\_xinwu@bjut.edu.cn
\IEEEcompsocthanksitem Guizeng.You,Chong.Zhao and Xinian.Guo are with  Beijing University of Technology.\protect\\
E-mail:18562310605@163.com,1935553270@qq.com,guoxinian@aliyun.com \protect\\
Mengfei.Dou is with College of Computer and Information Engineering, Henan Normal University.\protect\\ E-mail:doumf1110@163.com\protect\\
The work described in this paper was supported by Science Challenge Project,No.TZ2016003.}
}

\IEEEtitleabstractindextext{
\justifying
\begin{abstract}
  Two important characteristics of multi-objective evolutionary algorithms are distribution and convergency. As a classic multi-objective genetic algorithm, NSGA-II is widely used in multi-objective optimization fields. However, in  NSGA-II, the random population initialization and the strategy of population maintenance based on distance cannot maintain the distribution or convergency of the population well. To dispose these two deficiencies, this paper proposes an improved algorithm, OTNSGA-II II, which has a better performance on distribution and convergency. The new algorithm adopts orthogonal experiment, which selects individuals in manner of a new discontinuing non-dominated sorting and crowding distance, to produce the initial population. And a new pruning strategy based on clustering is proposed to self-adaptively prunes individuals with similar features and poor performance in non-dominated sorting and crowding distance,  or to individuals are far away from the Pareto Front  according to the degree of intra- class aggregation of clustering results. The new pruning strategy makes population to converge to the Pareto Front more easily and maintain the distribution of population. OTNSGA-II and NSGA-II are compared on various types of test functions to verify the improvement of OTNSGA-II in terms of distribution and convergency.
\end{abstract}

}

\maketitle

\IEEEdisplaynontitleabstractindextext

\IEEEpeerreviewmaketitle

\IEEEraisesectionheading{\section{Introduction}\label{sec:introduction}}

\IEEEPARstart{I}{n} real production and life, most of the actual problems require comprehensive optimization of multiple objectives, which are often highly complex and non-linear. Multi-Objective Evolutionary Algorithms (MOEAs) are very suitable for solving such problems and have become one of the mainstream algorithms for solving multi-objective optimization problems \cite{Ref1}. The purpose of the multi-objective evolutionary algorithm is to make the solution set approaches the Pareto ideal frontier of the problem, and the distribution is wide and uniform, which makes the distribution and convergency become the performance index  to evaluate algorithms \cite{Ref2}.Distribution and convergency are of great significance to solving multi-objective optimization problems. Well distribution can provide more reasonable and effective options for decision makers, and well convergency can solve practical problems more accurately. In recent years, some multi-objective evolutionary algorithms have been proposed on evolutionary computing. Among them, the most representatives are: Zitzler and Thiele proposed the SPEA \cite{Ref3}, Kim et al. proposed on the basis of SPEA2 \cite{Ref4}, Srinivas and Deb proposed non-dominated sorting genetic algorithm NSGA \cite{Ref5}, others on the basis of NSGA-II \cite{Ref6}, and \cite{Ref7} Pareto Envelope-based Selection Algorithm and PESA-II \cite{Ref8} proposed by Corne. NSGA-II is one of the most widely used multi-objective evolutionary algorithms. Its characteristic is the determination of individual fitness values are based on the Pareto dominance relationship and density information between individuals \cite{Ref9}. However, there are shortcomings of such a fitness calculation in improper maintenance of distribution and convergency. Wen Shihua et al. retained some representative individuals to improve NSGA-II according to the distance measurement method \cite{Ref10}. This method only considers the influence of crowding distance on the distribution of population, and does not take the perspective of the convergency and distribution problems of same kind of individuals caused by the existence of similar individuals and inferior individuals. And the influence of the distribution of initial populations on subsequent evolutionary processes is not considered. Aim at these two deficiencies of NSGA-II, this paper proposes an improved convergency and distribution retention strategy. Before the evolution, the multi-objective orthogonal experiment was used to initialize the population by using the non-dominated sorting and crowding distance evaluation to prevent the population from falling into local convergence or slow convergence due to random initialization, and to avoid the unevenness of individuals inside the initial population which leads to the lack of distribution. In the evolutionary process, the algorithm clusters each evolution, dynamically adjust the strength of the pruning within the class according to the similarity, and maintain the population by adaptively removing the appropriate number of in-class and non-dominated sorting and intra-class individuals with poor crowding distance and a small number of individuals far from the frontier face, speeding up the population convergence while maintaining the distribution. The comparison experiments between OTNSGA-II and NSGA-II show that the algorithm has good distribution and convergency.

\section{Related concepts}

The following describes the definition of multi-objective optimization problems from a rigorous mathematics perspective. The mathematical definition of the multi-objective optimization (MOP) problem that is commonly adopted and generally accepted in the field of multi-objective optimization is as follows:\newline
Definition ~\ref{1} (MOP): A general MOP consists of n decision variable parameters, k objective functions, and m constraints. The objective function, constraints, and decision variables are functional relationships. The optimization objectives are as follows:
\begin{align}
\label{1}
\begin{array}{*{20}{l}}
{Maximize\;\;\;\;y = f(x) = ({f_1}(x),{f_2}(x),...,{f_k}(x))}\\
{S.t.\;\;\;\;\;\;\;\;\;\;\;e(x) = ({e_1}(x),{e_2}(x),...,{e_m}(x)) \le 0}\\
{{\rm{among\;\;\;\;\;\;\;\;\;\;\;x}} = ({x_1},{x_2},...,{x_n}) \in X}\\
{\;\;\;\;\;\;\;\;\;\;\;\;\;\;\;\;\;\;\;\;\;y = ({y_1},{y_2},...,{y_k}) \in Y}
\end{array}
\end{align}

Where x represents the decision vector, y represents the target vector, X represents the decision space formed by the decision vector x, Y represents the target space formed by the target vector y, and the constraint  \begin{math}e(x)\le0\end{math} determines the feasible range of values of the decision vector. \newline
\indent When there are multiple objective functions, the concept of "optimal solution" produces new changes. When solving a multi-objective problem, it is actually seeking a set of equilibrium solutions instead of a single global optimal solution. The concept of this widely adopted optimal solution was proposed by Francis Ysidro Edgeworth early in 1881. Subsequently, the famous French economist and sociologist Vilfredo Pareto promoted the concept in 1896. From the perspective of economics, he transformed multiple incomparable objects into single indicators for optimal solution, involving the concept of multi-objectives. Pareto first proposed the concept of vector optimization, which is Pareto optimal that is used widely nowadays.\newline
\indent    The difficulty of MOP is that in most cases, the targets may be conflicting and the improvement of one target may cause the performance of other targets to be reduced. It is generally impossible to achieve multiple targets at the same time. Otherwise, it is not multi-objective optimization. The ultimate solution to the MOP can only be compromising and balancing among the objectives, so that the objective functions can be optimized as far as possible. Therefore, the optimal solution of MOP is essentially different from the optimal solution of single-objective optimization problem. In order to solve MOP correctly, the concept of solution must be defined.\newline
    Definition ~\ref{2} (feasible solution set): The feasible solution set $X_{f}$ is defined as a set of decision vectors x satisfying the constraint condition $e(x)$ in equation ~\ref{1}
\begin{equation}
\label{2}
x_f=\left\{x\in X\vert e\left(x\right)\leq0\right\}\;\;
\end{equation}
The expression of the target space corresponding to the feasible region of $X_{f}$ is:
\begin{equation}
\label{3}
Y_f=f\left(x_f\right)=Y_{x\in X}\left\{f\left(x\right)\right\}
\end{equation}
Equation ~\ref{3} represents that all xs in the feasible solution set Xf form a subspace in the target space through the optimization function mapping, and the decision vectors of the subspace belong to the feasible solution sets.\newline
\indent      For the minimization problem, it can be easily converted to the above maximization problem to solve.\newline
\indent      The feasible solution set of the single-objective optimization problem can determine the merit relationship and degree of the scheme through the unique objective function f. For the MOP problem, the situation is different, because in general, the decision vector in $X_{f}$ cannot be sorted completely, but only a certain index can be sorted, that is, partially sorted  \cite{Ref11}. To this end, define the mathematical relationshipbetween decision vectors $=$, $>$, $\geq$  as follows:\newline
Definition ~\ref{4}: For vector u, v:
      \begin{equation}
      \label{4}
      \begin{array}{l}u=v:\;if\;and\;only\;if\;\forall_i\in\left\{1,2,...k\right\},u_i=v_i\\u\geq v:\;if\;and\;only\;if\;\forall_i\in\left\{1,2,...k\right\},u_i\geq v_i\\u>v:if\;and\;only\;if\;\forall_i\in\left\{1,2,...,k\right\},u_i>v_i\end{array}
      \end{equation}
     Similarly, the following mathematical relations can be defined: $<$,$\leq$.\newline
     \indent Then define the mathematical relationship between the decision vectors:\newline
     Definition ~\ref{5} (Pareto win): For decision vectors a, b:

      \begin{equation}
      \label{5}
      \begin{array}{l}
      a \succ b,if\;\;and\;\;only\;if\;\; f\left( a \right) > f\left( b \right)\\
      a \succeq b,if\;\; and\;\;only\;if\;\; f\left( a \right) \ge f\left( b \right)\\
      a \sim b,if\;\; and\;\;only\;if\;\; f\left( a \right) \ngeq f\left( b \right),f\left( a \right) \nleq f\left( b \right)
      \end{array}
      \end{equation}

\indent If the vector a, b has a Pareto superior relationship, the values of all the objective function of one vector are greater than the values corresponding to the other vector.  Now define the concept of the Pareto optimal solution:\newline
Definition ~\ref{6} (Pareto optimal solution): The decision vector $x\in X_{f}$ is called the Pareto optimal solution if and only if:
  \begin{equation}
  \label{6}
  \not\ni a\in X_f:f\left(a\right)\succ f\left(x\right)
  \end{equation}
The Pareto optimal solution is also known as non-inferior solutions or Non-dominated solutions. If the decision vector x is a Pareto optimal solution, this means that x cannot improve any of the objective functions without weakening at least one other objective function. For a Pareto optimal solution in a given target space Y, its corresponding point in the decision space is called efficient or non-inferior. One point in $X_{f}$ is efficient if and only if its image is non-dominant in Y.
\indent Different from the Single-objective Optimization Problem, there is usually no unique optimal solution in MOP problem, but a Pareto optimal solution set. The concept of Pareto optimal solution set is as follows:\newline
Definition ~\ref{7} (Pareto optimal solution set): For a given MOP problem, the Pareto optimal solution set A is defined as:
     \begin{equation}
     \label{7}
     A=\left\{x\in X_f\vert\nexists\;x'\in X_f,f\left(x'\right)\succ f\left(x\right)\right\}
     \end{equation}
\indent In most cases, in the multi-objective optimization problem, only the Pareto optimal solution exists \cite{Ref12}, and the optimal solution similar to the single-objective optimization does not exist. Multi-objective optimization problems, mostly with many Pareto optimal solutions, its Pareto optimal solution is only an acceptable non-inferior solution or satisfactory solution. If there are so-called optimal solutions for a multi-objective optimization problem, they must be a Pareto optimal solution consisting of only these optimal solutions and no other solutions. Therefore, the Pareto optimal solution is A reasonable solution set for multi-objective optimization problems \cite{Ref13} \cite{Ref14}. For the practical application problem, one or more solutions must be selected from the Pareto optimal solution of the multi-objective optimization problem as the optimal solution for the multi-objective optimization problem according to the degree of understanding of the problem and the personal preference of the decision-makers \cite{Ref15} \cite{Ref16}. Therefore, the first step and key to solving the multi-objective optimization problem is to find as many Pareto optimal solutions as possible.\\ Two goals for multi-objective optimization are as follows:\\
     (1) Find a set of solutions that are as close as possible to the Pareto optimal domain.\\
     (2) Find a set of solutions that are as different as possible and evenly distributed.\\
\indent The first goal is necessary in any optimization work. It is not advisable to converge to a set of solutions that are not close to the true Pareto optimal solution set. Only when the solution converges to approximate true optimal solutions can we guarantee their approximate optimal properties. In addition to converge to the approximate optimal domain, the solutions obtained must also be distributed as evenly as possible throughout the Pareto optimal domain. Only by finding a set of solutions that are sufficiently diverse and covering as much as possible can each of them include a number of different effective solutions for decision makers to choose.

\section{Orthogonal Initial Population}
\subsection{The problem of NSGA-II population initialization}
\indent In the NSGA-II method, the initial population is formed by using the random initialization method, that is, the value of the solution is randomly selected within the range of the values of the independent variables of each dimension, thereby forming a new individual, which will result in the distribution of randomly generated individual sets is not uniform, and most of the individuals are not close to the frontier, making the early evolutionary exploration easy to fall into the local optimum, which is not conducive to converge to the Pareto frontier, and also causes the lack of distribution. The randomly initialized population is shown in Figure ~\ref{fig:random_initial}:

\begin{figure}[!h]
\centering

\includegraphics[width=2.5in]{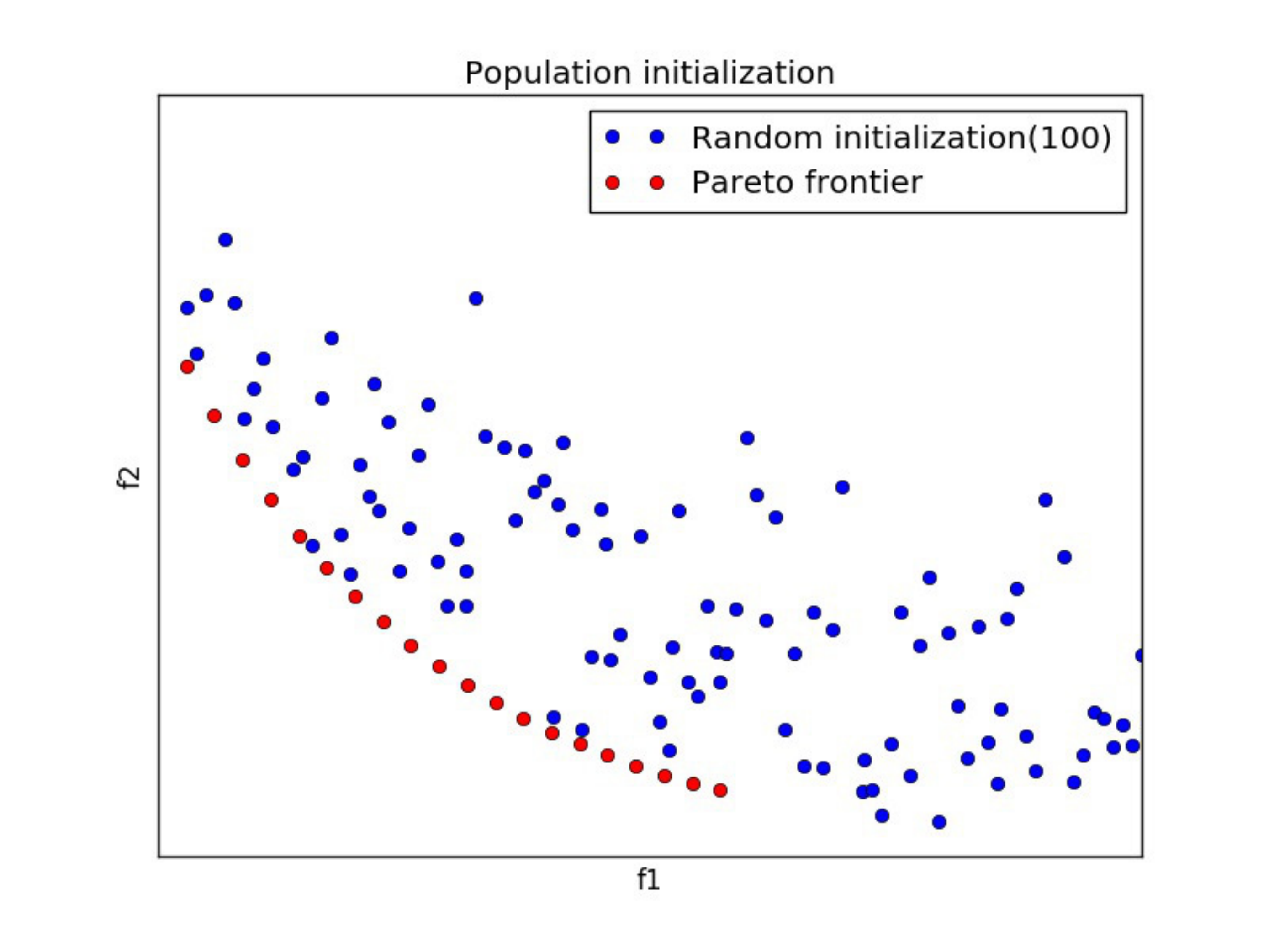}

\caption {Random initial population.}
\label{fig:random_initial}   
\end{figure}

\indent In Figure ~\ref{fig:random_initial}, we can see that the initial population distribution generated by this random initialization method is not uniform enough, some individuals are too concentrated, some individuals are too separated; and most individuals are far from Pareto ideal frontier, which is not conducive to the maintenance of convergency and distribution in the subsequent evolution process inevitably.

\subsection{Method for setting initial population in fault multi-objective}
Orthogonal experimental design, as a method that can solve multi-factor and multi-level experimental problems effectively \cite{Ref17}, uses orthogonal tables to arrange a few experiments to find the best or better experimental conditions \cite{Ref18}\cite{Ref19}\cite{Ref20}, so it is widely used  in solving the problem of optimization \cite{Ref20}. Suppose an experimental system has F factors, each of each with Q levels. If a comprehensive combination test is performed, a  experiments are required. While using orthogonal table $L_{M}(Q^{F})$ , just choose $M$ combinations to test. Where $L_{M}(Q^{F})$ represents an orthogonal table with $F$ factors and $Q$ levels, $L$ represents the Latin square, $M$ represents the number of horizontal combinations, there are $M$ rows, each row represents a horizontal combination, and $M$ is generally much smaller than \cite{Ref21}. For the four-factor and three-level problems, if all the horizontal combinations are fully tested, a total of $3^4 = 81$ tests are required; and if an orthogonal table is used, such as formula ~\ref{8}, only 9 tests are required, which can greatly reduce the number of experiments.

\begin{equation}
\label{8}
L_9\left(3^4\right)=\begin{bmatrix}1&1&1&1\\1&2&2&2\\1&3&3&3\\2&1&2&3\\2&2&3&1\\2&3&1&2\\3&1&3&2\\3&2&1&3\\3&3&2&1\end{bmatrix}
\end{equation}
For convenience of representation, it is noted that $L_{M}(Q^{F})=[a_{i,j}]_{M \times F}$ , where $a_{ij}$ represents the horizontal value of the $j_{th}$ factor of the ith combination , if $j_{th}$ satisfies $j=1,2,(Q^{2}-1)\div(Q-1)+1,(Q^{3}-1)\div(Q-1)+1,...(Q^{j}-1)\div(Q-1)+1$ , the $j_{th}$ column is called the basic column, other columns are non-basic columns. Orthogonal table creation process is as follows \cite{Ref22}\cite{Ref23}:

\begin{algorithm}[!h]
  \caption{Construction of Orthogonal Array Algorithm}
  \label{code:create_Orthogonal_table}
  \begin{algorithmic}[1]
  \State{Step 1: Calculate the minimum value J that satisfies} \State{$\left(Q^J-1\right)/\left(Q-1\right)\geq F$}
  \State{Step 2:}
  \If {$\left(Q^J-1\right)/\left(Q-1\right)=F$}
  \State{$F^{'} =F$}
  \Else
  \State{$F^{'} =\left(Q^J-1\right)/\left(Q-1\right)$}
  \EndIf
  \State{Step 3: Construct the basic columns:}
  \For{$k=1$ to $J$ do}
  \State $j=\frac{Q^{k-1}}{Q-1}+1$
  \For{$i=1$ to $i<Q^{J}$ do}
  \State{$a_{i,j}=\left\lfloor \frac{i-1}{Q^{j-k}}\right\rfloor$ mod  Q}
  \EndFor
  \EndFor
  \State{Step 4:Construct a non-basic column:}
  \For{$k=2$ to $J$ do}
  \State $j=\frac{Q^{k-1}}{Q-1}+1$
  \For{$s=1$ to $j-1$ do}
  \For{$t=1$ to $Q-1$ do}
  \State {$a_{j+\left(s-1\right)/\left(q-1\right)+t}=\left(a_s\times t+a_j\right)$mod Q}
  \EndFor
  \EndFor
  \EndFor
  \State{Step5: If $i$ and $j$ satisfy$ 1\leq i\leq M $ and $ 1 \leq j\leq F^{'} $, execute $ a_{i,j} =a_{i,j}+1$}
  \State{Step6: Delete the last $ F^{'}-F $columns of the orthogonal table $L_{M}(Q^{J})$ to get $L_{M}(Q^{J})$}
  \end{algorithmic}
\end{algorithm}
   
In order to make the population converge quickly to the Pareto frontier and distribute evenly, the initial population should be covered as much as possible to cover the entire feasible domain and close to the Pareto ideal frontier in the population initialization process of the multi-objective optimization problem. The multi-objective orthogonal experiment of fault is a multi-factor and multi-level mathematical experiment method from the perspective of Pareto multi-objective optimization. According to the orthogonal table, some representative points with uniform dispersion and uniformity are selected from the comprehensive experiment to carry out experiments. At the same time, for multi-objective problems, it is necessary to calculate the non-dominated sorting layer according to the relationship between multiple targets, and select a part of the individuals in the front layer as an alternative set. And in this alternative set, the method of evaluating the individual based on the non-dominated sorting stratification and the crowding distance is selected to better adapt to the initial population scene in which the multiple targets are mutually restrained while maintaining the convergence and distribution of the population. . The good distribution and convergency of the initial population ensure the diversity of the population and the richer model and close to the optimal solution set, so that the algorithm can converge at a faster speed in the global scope and has a uniform distribution and a wide coverage. When using the fault multi-objective orthogonal experiment method to initialize the population, if the feasible solution space $[l, u]$ is large, in order to improve the search efficiency and precision, the feasible solution space $[l, u]$ is divided into $S$ subspaces (For the segmentation method, see Algorithm ~\ref{code:subspace_segmentation} in this section), then construct the orthogonal table $L_{M}(Q^{F})$. According to the orthogonal table, for each subspace, the proposed SOC operator is used for cross operation, and a discretized initial individual set is generated, so that the candidate set is selected therefrom and the initial population is composed by selection.

 \begin{algorithm}[!h]
  \caption{Subspace Segmentation}
  \begin{algorithmic}[1]
  \State{Step 1: Select the $s_{th}$ dimension of the feasible solution space, where s satisfies the following formula:}
  \begin{equation}
  u_s-l_s=\mbox{\large$max$}_{1\leq i\leq N}\left\{u_i-l_i\right\}
  \end{equation}
  \State{Step 2: Split the feasible solution space [l, u] into S subspaces at the $s_{th}$ dimension[l(1),u(1)],[l(2),u(2)],...,[l(S),u(S)].}
  \begin{align}
  \left\{\begin{array}{l}l\left(i\right)=l+\left(i-1\right)\left({\textstyle\frac{u_s-l_s}S}\right)I_s\\u\left(i\right)=u-\left(S-1\right)\left(\frac{u_s-l_s}S\right)I_s\end{array}\right.
  ,i=1,2,...,S \\
  \ \notag
   where I_S={\left[c_{1,j}\right]}_{1\times N},C_{1,j}=\left\{\begin{array}{l}1,\;j=s\\0,\;j=0\end{array}\right.
  \end{align}
  \label{code:subspace_segmentation}
  \end{algorithmic}
\end{algorithm}

\begin{figure}[!h]
\centering

  \includegraphics[width=2.5in]{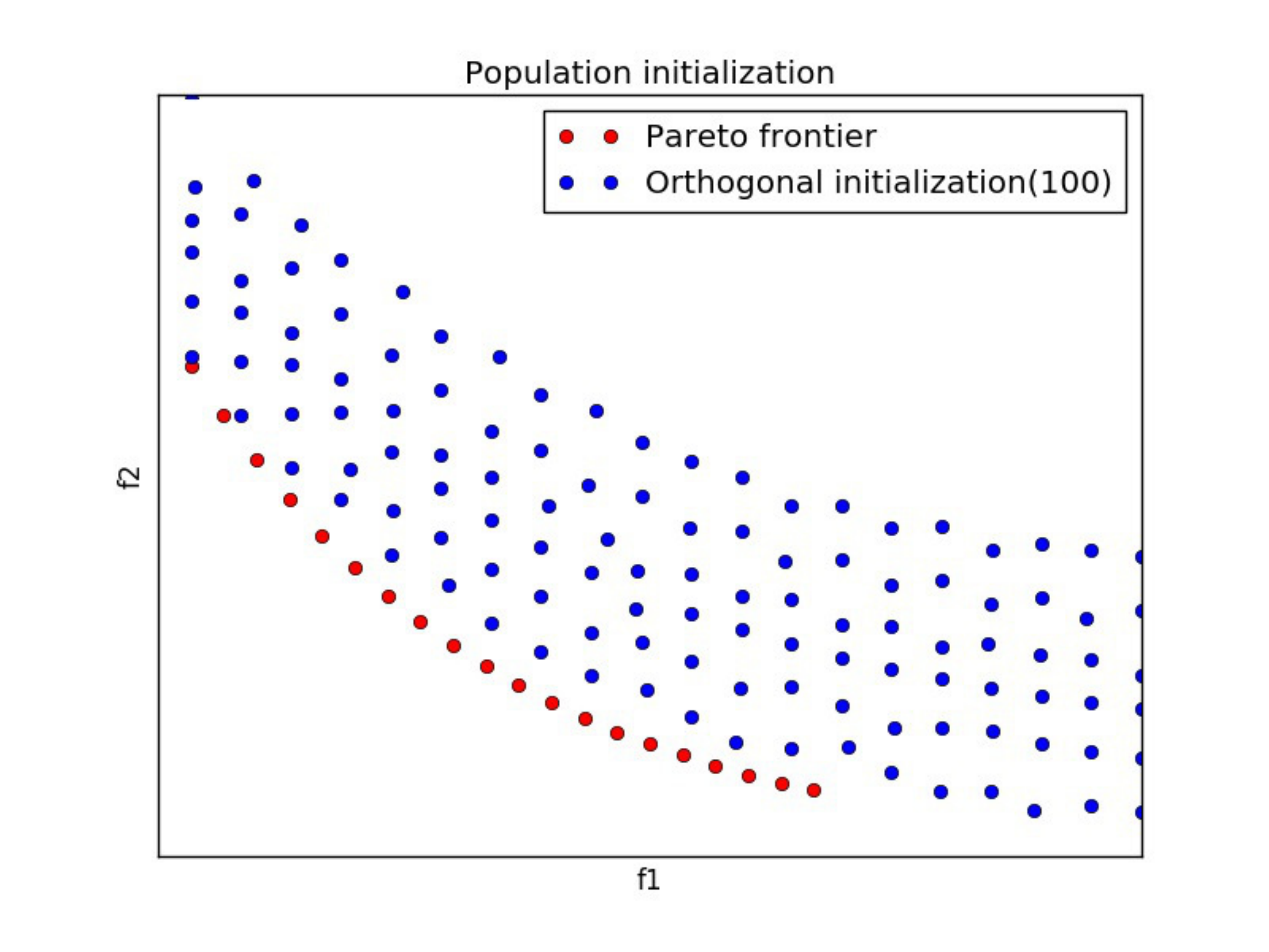}

\caption{Initialization of multi-objective orthogonal experimental population}
\label{fig:orthogonal_initialization}      
\end{figure}

\begin{algorithm}[!h]
\caption{ self-adaptive orthogonal crossover (SOC)}
\label{code:Self-adaptive_Orthogonal_Algorithm}
\begin{algorithmic}[1]
   \State{Step 1:Let $P_{1} = ( P_{1,1}, P_{1,2} ,..., P_{1,N} ) , P2 = (P_{1,1}, P_{1,2} ,..., P_{1,N} )$ be the two parent individuals involved in the cross operation. The feasible solution space determined by $P_1$ and $P_2$ is $[lparent, uparent]$. Then the ith dimension in the space$[lparent, uparent]$ is discretized into $Q$ levels, namely $\beta_{i1}, \beta_{i2},..., \beta_{iQ}, i\in\{1,2,...,N\}, and \beta_{i} =( \beta_{i1}, \beta_{i2} ,..., \beta_{iQ} )$ , where  :} 
  \begin{equation}
  {\beta _{ij}}\left\{ {\begin{array}{*{20}{l}}
  {\min ({p_{1,i}},{p_{2,i}}),j = 1}\\
  {\min ({p_{1,i}},{p_{2,i}}) + (j - 1)\left( {\frac{{\left| {{p_{1,j}} - {p_{2,j}}} \right|}}{{Q - 1}}} \right)}\\
  {\max ({p_{1,i}},{p_{2,i}}),j = Q}
  \end{array}} \right.,2 \le j \le Q - 1
  \end{equation}
  \State{Step 2:  Let the vector $k = [k_{1}, k_{2} , ..., k_{t} ]$ satisfies: $k_{i}\in J and 1 \le k_{1} < k_{2} < ... < k_{t} \le N , j = 1, 2, ..., t$, where set $J$ is the set of positions where the components of low similarity in the parental individuals $p_{1}$ and $p_{2}$ are located, and $t$ is the number of components of low similarity in $p_{1}$ and $p_{2}$, and $\theta 0$ is a given positive real number close to zero. The vector $k$ holds the position of the component with low similarity in $P_{1}, P_{2}$, that is, the position at which the factor is divided. Let $x$ be any individual in $P_{1}, P_{2}$, divide the individual $x = (x_{1}, x_{2},..., x_{N} )$ into $t$ parts, each of which represents a factor of individual $x$:}
  \begin{equation}
  \left\{ {\begin{array}{*{20}{l}}
  {{f_1} = ({x_1},...,{x_{k1}})}\\
  {{f_2} = ({x_{k1 + 1}},...,{x_{k2}})}\\
  \ldots \\
  {{f_t} = ({x_{t - 1 + 1}},...,{x_N})}
  \end{array}} \right.
  \label{12}
  \end{equation}

  \State{Let $k_{0} = 0$, then the $Q$ levels of the $i_{th}$ factor $f_{i}$ can be expressed as:}
  \begin{equation}
  \left\{ {\begin{array}{*{20}{l}}
  {{f_i}(1) = ({\beta _{ki - 1 + 1,1}},{\beta _{ki - 1 + 2,1}},...,{\beta _{ki,1}})}\\
  {{f_i}(2) = ({\beta _{ki - 1 + 1,2}},{\beta _{ki - 1 + 2,2}},...,{\beta _{ki,2}})}\\
  \ldots \\
  {{f_i}(Q) = ({\beta _{ki - 1 + 1,Q}},{\beta _{ki - 1 + 2,Q}},...,{\beta _{ki,Q}})}
  \end{array}} \right.
  \label{13}
  \end{equation}
  \State{Step 3:  According to the algorithm ~\ref{code:create_Orthogonal_table}, construct the orthogonal table $L_{M}(Q^{F})$=$[b_{i,j}]_{M\times F}$, where $F = t$. Based on the orthogonal table $L_{M}(Q^{F})$, orthogonal experimental design is performed for the t factors determined in equation \eqref{12} and the $Q$ levels corresponding to each factor determined in equation \eqref{13}, and $M$ child generation individuals will be generated:}
  \begin{equation}
  \left\{ {\begin{array}{*{20}{l}}
  {({f_1}({b_{1,1}}),{f_2}({b_{1,2}}),...,{f_F}({b_{1,F}}))}\\
  {({f_1}({b_{2,1}}),{f_2}({b_{2,2}}),...,{f_F}({b_{2,F}}))}\\
  {...}\\
  {({f_1}({b_{M,1}}),{f_1}({b_{M,2}}),...,{f_F}({b_{M,F}}))}
  \end{array}} \right.
  \end{equation}
  \end{algorithmic}
\end{algorithm}
The OTNSGA-II design initializes the population by multi-objective orthogonal experiments by using faults and non-dominated sorting and crowding distances for individual selection. Using this method of tomographic multi-objective orthogonal experiments, the population is initialized before evolution, so that the initialized individuals can be evenly distributed at different positions close to the Pareto front, preventing the population from falling into local convergence or convergence is too slow due to random initialization, and avoiding the uneven distribution of the initial population individuals leads to the loss of distribution and thus is not conducive to extending to the entire Pareto front as much as possible. The population initialized by the fault multi-objective orthogonal experiment is shown in Figure ~\ref{fig:orthogonal_initialization}.   \\
\indent In Figure ~\ref{fig:orthogonal_initialization}, we can see that the initial population of the fault multi-objective orthogonal experiment is evenly distributed, and most individuals are close to the Pareto ideal frontier, which not only maintains good initial distribution and convergence, but also lays the foundation for the continued maintenance of the above two excellent characteristics in the subsequent evolution process.

\begin{algorithm}[!h]
  \caption{Orthogonal initialization population}
  \label{code:Orthogonal_Initial_Population}
  \begin{algorithmic}[1]
  \State{Step 1: Divide the feasible solution space $[L, U]$ in each dimension of the optimization problem into $S$ subspaces, and the segmentation method is shown in Algorithm ~\ref{code:subspace_segmentation}}
  \State{Step 2: Take the horizontal number of the orthogonal table as $Q_{0}$, and cross-operate each subspace based on the SOC operator (algorithm ~\ref{code:Self-adaptive_Orthogonal_Algorithm}) to generate a new discretized population $P$, where $Q_{0}$ is the discretized horizontal number;}
  \State{Step 3: Calculate the value fit of the target of each individual in the population $P$ in the problem to be optimized, expressed as the value of the $i_{th}$ individual on the $i_{th}$ objective function;}
  \State{Step 4:Perform calculation of non-dominated sorting stratification according to the superiority and inferior relationship between each target value fit for each individual in the population $P$, and mark the level $c_{i}$ of each individual, which is represented as the level at which the $i_{th}$ individual is located;}
  \State{Step 5:According to the hierarchical ranking, the individual layers are sequentially acquired from the first layer into another set until the number of individuals in the set is greater than or equal to $4*n$.( If the individual number of population $P$ is less than $4*n$, take the individual collection of the population.) and the number of different layers in the set is $m$. This set is an alternative set, where $n$ is the number of individuals in the population;}
  \State{Step 6:Calculate the congestion distance $d_{i}$ as the value of the $i_{th}$ individual congestion distance for each individual in the candidate set by using the candidate set as the ensemble space;}
  \State{Step 7:In the alternative set, the non-dominated sorting level $c_{i}$ and the crowding distance $d_{i}$ of each individual are taken as two targets, and the non-dominated relationship between the individuals is evaluated to select the optimal first $n$ individuals to generate the initial population P0.}
  \end{algorithmic}
\end{algorithm}
From this algorithm flow that maintains the convergency and distribution of the initial population, it can be seen that: In the process of designing the fault multi-objective orthogonal experiment to initialize the population, according to the multi-factor and multi-level characteristics, the distributed population with uniform distribution and neatness is generated, which ensures excellent initial distribution. Selecting the individuals with higher non-dominated stratification in the discretized population into the candidate set maintains the characteristics of these individuals close to the Pareto ideal frontier and improves the initial convergency performance of the population. In the alternative set, according to the non-dominated sorting stratification and the crowding distance calculation, the better individuals are selected to form the initial population, so that the maintenance of the population distribution is comprehensively considered while ensuring the proximity to the ideal frontier. However, the way in which the algorithm NSGA-II randomly initializes the population leads to uneven distribution of individuals and most individuals are far away from the ideal frontier, which is not conducive to the convergency and distribution.

\section{Adaptive Clustering Pruning Strategy}

\subsection{Problems in the calculation of crowding distance in NSGA-II}
\begin{figure}[!h]
\centering

  \includegraphics[width=2.5in]{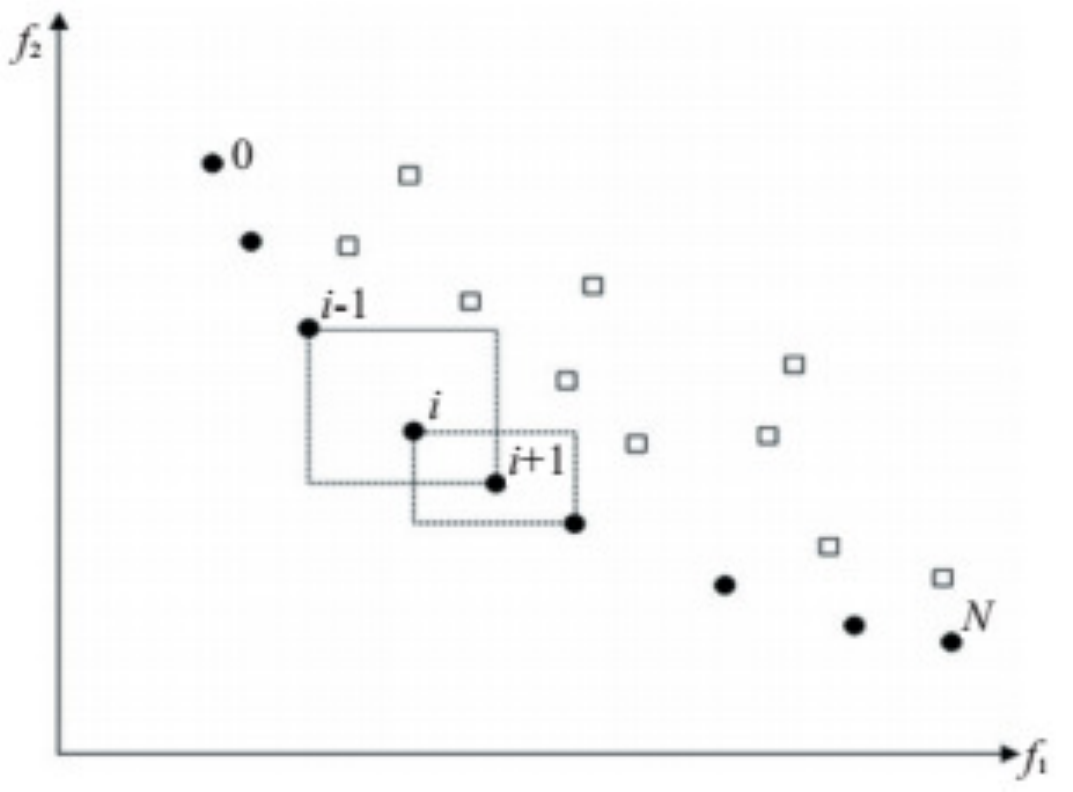}

\caption{Schematic diagram of congestion distance calculation}
\label{fig:Crowding distance calculation}       
\end{figure}

NSGA-II ranks each individual hierarchically by the non-dominant relationship of each target between individuals and takes the number of layers as the quantitative fitness value of the individual, and calculates the crowding distance between individuals in the population. Then each individual is ranked according to the number of layers in which the individual is located and the calculated crowding distance [24], and selects a certain number of individuals from the front to form the next generation of new populations. The crowding distance of each individual is the sum of the distance differences between the individual and the two individuals adjacent to each other in each target dimension [25] [26]. As shown in Fig. 4-1, two targets f1 and f2 are provided, and the crowding distance of the individual i is the sum of the length and the width of the dotted quadrilateral in the figure. Let d[i]distance be the crowding distance of the individual i, and d[i].m is the function value of the individual i on the target m, then the crowding distance of the individual $i$ in Figure~\ref{fig:Crowding distance calculation} is:

\begin{equation}
\begin{split}
\label{16}
d{\left[ i \right]_{distance}} = \left( {d{{\left[ {i + 1} \right]}_{.f1}} - d{{\left[ {i - 1} \right]}_{.f1}}} \right) \\
+ \left( {d{{\left[ {i + 1} \right]}_{.f2}} - d{{\left[ {i - 1} \right]}_{.f2}}} \right)
\end{split}
\end{equation}
In general, when there are $r$ targets, the crowding distance of individual $i$ is:

\begin{equation}
d{\left[ i \right]_{distance}} = \sum\limits_{k = 1}^r {\left( {d{{\left[ {i + 1} \right]}_{.fk}} - d{{\left[ {i - 1} \right]}_{.fk}}} \right)}
\end{equation}
The method of calculating the crowding distance of an individual is as shown in Algorithm ~\ref{code:crowding-distance-assignment}.

\begin{algorithm}[!h]
  \caption{crowding-distance-assignment}
  \label{code:crowding-distance-assignment}
  \begin{algorithmic}[1]
  \State{$ {\L}= \left| I \right|$}
  \State{for each i,set $I[i]_{distance}=0$}
  \State{for each objective m}
  \State{I=sort(I,m)}
  \State{$I[l]_{distance}=I[l]_{distance}=\propto$}
  \State{for $i =2$ to $({\L}-1)$}
  \State{$I[i]_{distance}=I[i]_{distance}+(I[i+1]_{.m}-I[i-1]_{.m})\div(
f_m^{\max } - f_m^{\min })$}
  \end{algorithmic}
\end{algorithm}

However, using the above crowding distance calculation method to select individuals may result in some individuals with good distribution to be eliminated, while others with poor distribution may be retained, which will lead to the concentration of individuals in some areas, and the sparseness of others in other areas \cite{Ref27}\cite{Ref28}, which is not conducive to the maintenance of population distribution; and a small number of individuals away from the frontier may also be retained, which will reduce the possibility that the next generation of new populations will continue to evolve to produce new superior individuals, which is not conducive to populations.

\begin{figure}[!h]

  \includegraphics[width=2.5in]{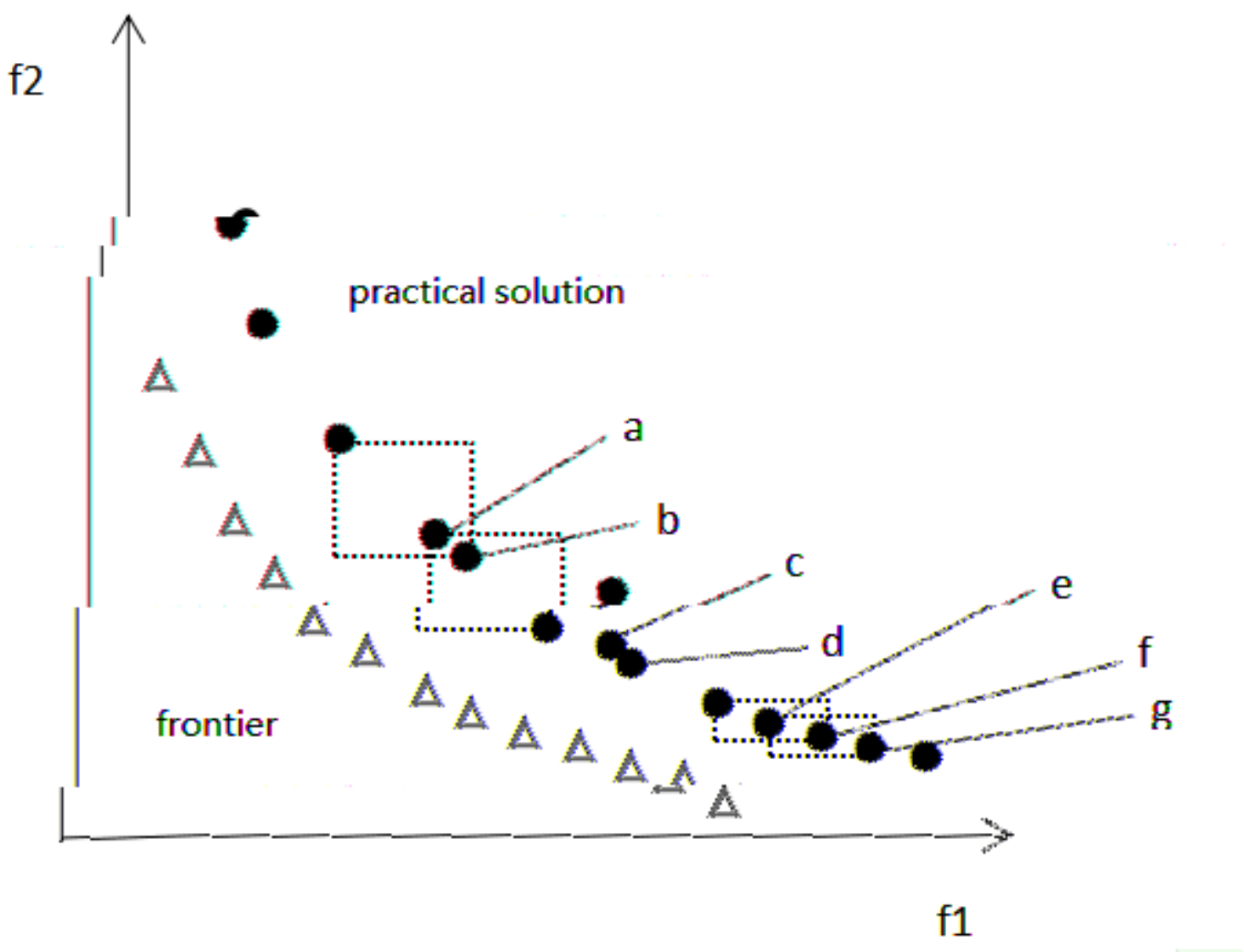}

\caption{A generation of evolutionary solution sets.}
\label{fig:individual}       
\end{figure}
As shown in Figure~\ref{fig:individual}, for a generation of solution sets in the evolution process, individuals $a$ and $b$ have similar characteristics, showing that the two positions are close and far from other points. Using NSGA-II method to calculate the crowding degree of individuals $a$ and $b$, the values are relatively close and relatively large, indicating that individuals $a$ and $b$ are in the same set of similar features, then the points in this collection are likely to be eliminated or retained at the same time, resulting in individuals in the region being too sparse or too dense, and the ideal situation is to retain a better performing individual in the set as a representative point with the feature set. The same is true for individuals $c$ and $d$. Individuals $e$, $f$, and $g$ are more evenly distributed, but the crowding distance obtained by the NSGA-II method will be small, and it is likely to be eliminated during the selection process, resulting in the loss of individuals in this area and the ideal situation is to retain one or two excellent representative points to maintain the distribution characteristics of similar collections of this feature. It can be seen that the calculation method of the crowding distance will affect the retention of the population distribution.

\begin{figure}[!h]

  \includegraphics[width=2.5in]{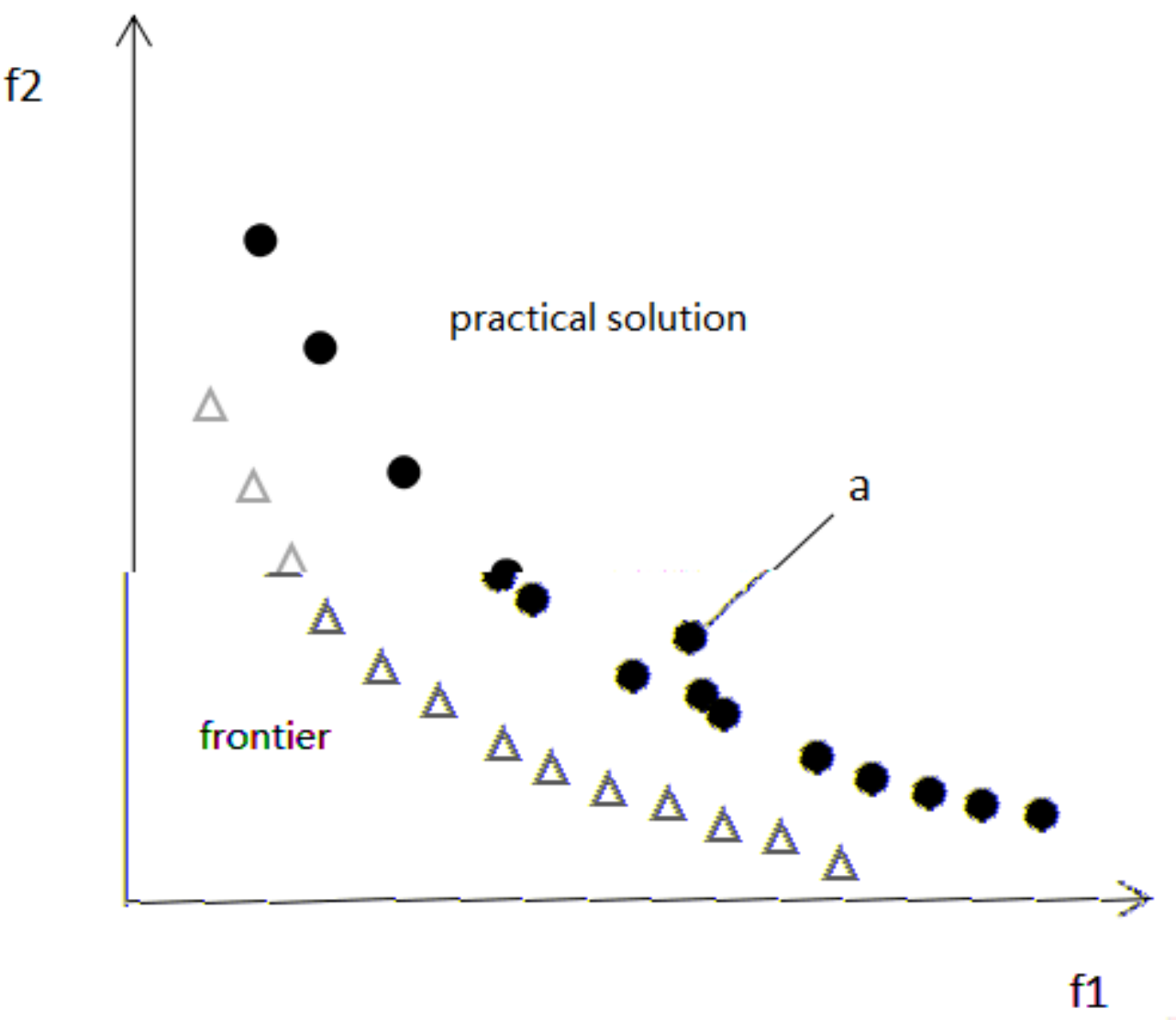}

\caption{Schematic diagram of a generation of evolutionary solution set}
\label{fig:A generation of evolutionary solution sets}       
\end{figure}

As shown in Figure ~\ref{fig:A generation of evolutionary solution sets}, according to the calculation method of NSGA-II, individual $a$ will enter the next generation for subsequent evolution. However, for the characteristic individual $a$, unlike other individuals who are far from the Pareto front, the individual $a$ tends to produce some poorly performing individuals in the next generation of evolution, which is detrimental to the rapid convergence of the population.It can be seen that this calculation method also affects the convergency performance of the population. 

\begin{figure}[!h]

  \includegraphics[width=2.5in]{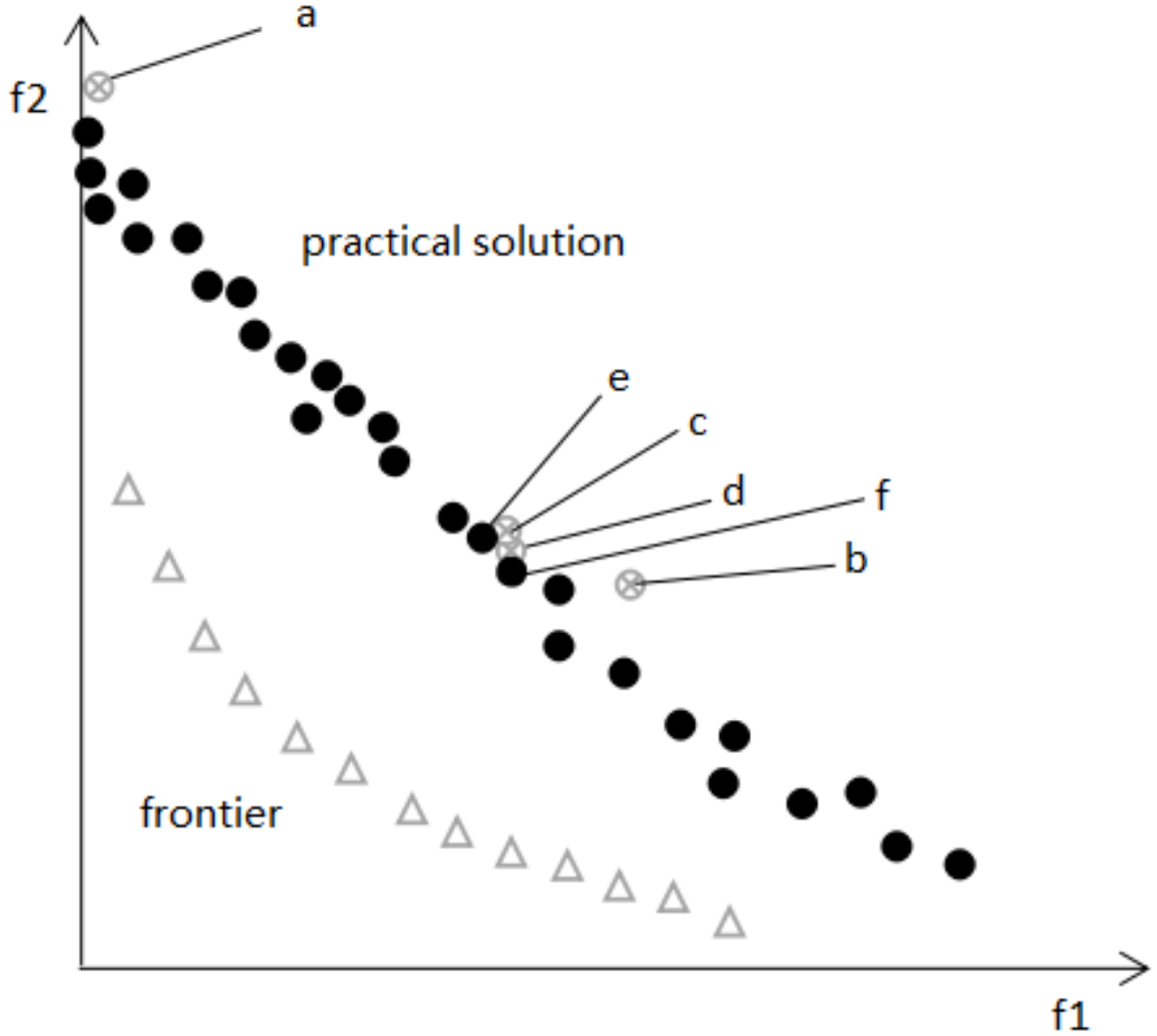}

\caption{Schematic diagram of a generation of evolutionary solution sets}
\label{fig:A generation of evolutionary solution sets_1}       
\end{figure}

A concrete example is further explained in Figure~\ref{fig:A generation of evolutionary solution sets_1}: in this figure, because the rankings of individuals $a$ and $b$ in the non-dominated sorting are ranked last and away from the Pareto frontier,  its existence is not conducive to the generation of good individuals in the subsequent evolution process, resulting in the entire population converge too slowly; although individuals $c$, $d$, $e$, and $f$ are in the same sorting layer,  because their characteristics are the same and the crowding distance is too low, they belong to the same set of individuals with similar characteristics, which may result in being eliminated or retained at the same time. If it is eliminated at the same time, the point set of this segment is empty, which is not conducive to the expansion to the entire frontier of Pareto; If too many individuals with similar characteristics are retained at the same time, it is not conducive to the combination of new individuals in the process of evolution, which leads to the excessive distribution of these individuals, which hinders the distribution maintenance of the solution set.

\subsection{Improved pruning strategy}
In the method of the algorithm OTNSGA-II, the individuals in the population are first clustered, and the appropriate number of non-dominated sorting and individuals with poor crowding distance are adaptively trimmed according to the similarity within the class. In this way, it is ensured that some points with poor ranking and congestion distance are removed from each set with similar features, and most of the non-dominated sorting points and the representative points with better crowding distance are retained, so that the point sets with the same characteristics will not be the same which is not conducive to further expansion of the solution set, and it is easier to combine excellent individuals; filtering a few point sets far away from the front surface is beneficial to the next generation of evolution to quickly converge to the Pareto front. In this way, the distribution and convergency of the population are maintained from the above two aspects. There are three basic issues that need to be addressed in this way: (1) the calculation of the average similarity within the class; (2) the setting of the number of individuals within the class; (3) the choice of the pruned individual. \\  
The calculation of the average similarity within the class: This paper uses K-means to cluster the population of each iteration result, calculate the similarity for any two bodies in the class, and then calculate the average of all two body similarities as the intra-class average similarity, the formula is as follows:
\begin{equation}
\label{similarity}
{\sum\limits_{i=1}^{M}}{\sum\limits_{j=i+1}^{M}p_{ij}}{/(M(M-1)/2)}
\end{equation}
Where $i$, $j$ represent an individual, $M$ is the number of individuals within the class, and $p_{ij}$ represents the similarity between two individuals $i$, $j$.   
Set the number of individuals in the reserved class: $1- \delta* the\quad average\quad similarity\quad within\quad each\quad class$, and the average similarity within the class is 0~1 (In practice, it is usually 0~0.8, because there are few cases where the average similarity within the class is 1, that is, all the individuals in the class repeat.) In the case of guaranteeing the main flow of the algorithm, adaptively trimming a small number of inferior individuals based on the similarity within the class, retaining most of the representative points with the similar feature set , the formula is as follows:
\begin{equation}
\label{number}
1- \delta*p_k
\end{equation}
 $p_k$refers to the intra-class average similarity in the above formula ~\ref{similarity}. Due to the characteristics of the solution set distribution of different types of functions and the different dimensions of each target, a large number of experiments have found that the appropriate adaptive pruning parameters $\delta$ are 0.12-0.15. The parameter should not be too large or too small. If it is too large, it will trim out too many individuals so that some excellent individuals can be trimmed together; if too small, the effect of removing inferior individuals with higher similarity is not obvious.   \\
\indent The method of selecting the pruned individuals: in the same class, retaining the above number of individuals based on a comparison rule that considers the non-dominated sorting firstly and considers the crowding distance secondly, that is, the selection is based on the non-dominated sorting hierarchical ranking and the value of the crowding distance. For two Individuals, if the non-dominated sorting hierarchy is different, individuals with larger levels are selected; if the non-dominated sorting level is the same, individuals with larger crowding distances are selected.
    
\begin{figure}[!h]

  \includegraphics[width=2.5in]{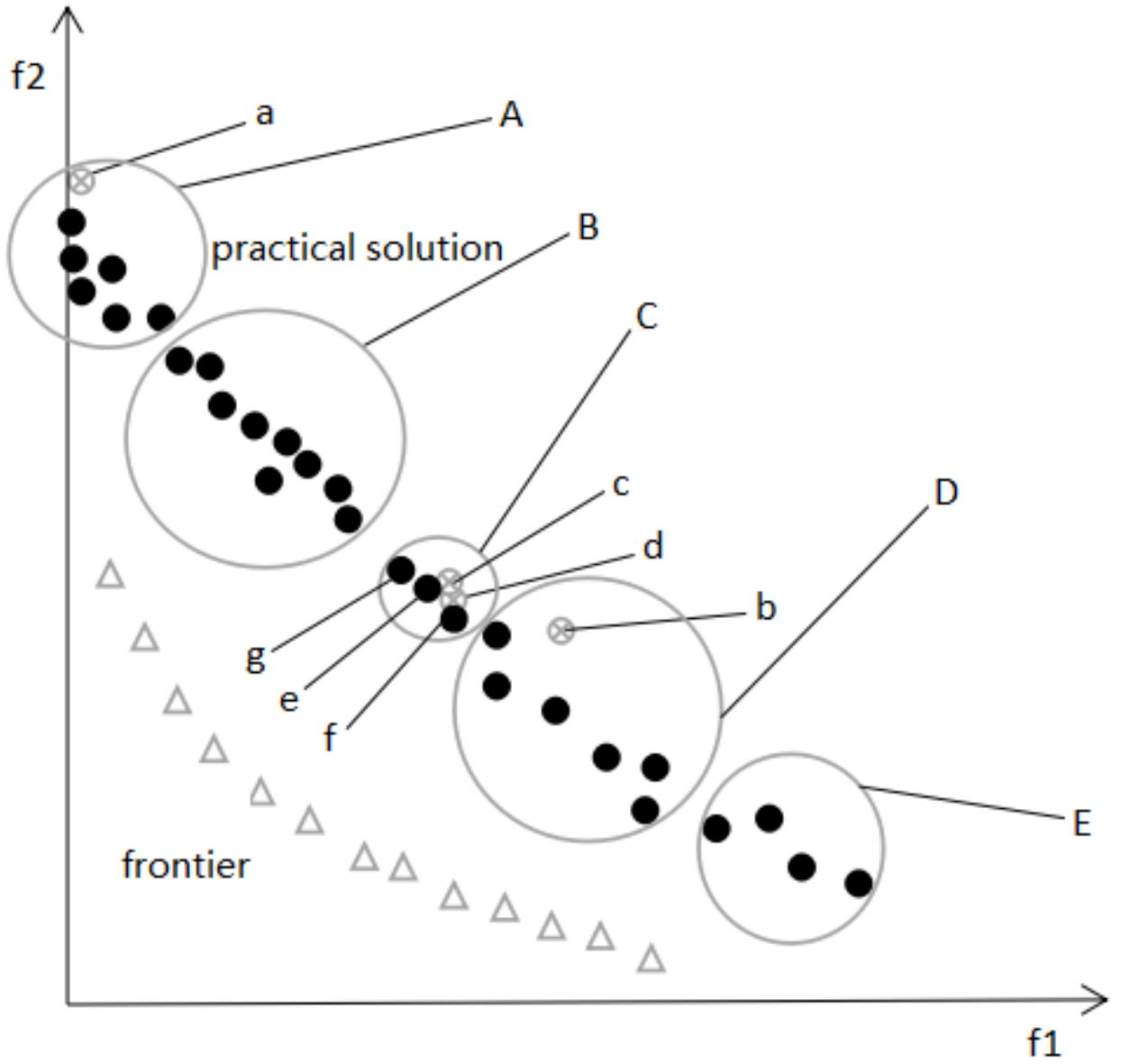}
\caption{Schematic diagram of a generation of evolutionary solution sets}
\label{fig:A generation of evolutionary solution sets_3}      
\end{figure}

 For example, as shown in Figure ~\ref{fig:A generation of evolutionary solution sets_3}, for a certain generation of the solution set in the evolution process, taking the coordinate values in each dimension of the individual and the non-dominated hierarchical ranking and the crowding distance as features, K-means clustering is performed on the solution set and is divided into five categories (areas in the five graphs $A$, $B$, $C$, $D$, and $E$). Through the calculation of the average similarity within the class, some individuals with the same non-dominated hierarchical ranking and poor crowding distance in the same category are filtered out. The higher the intra-class similarity, the more individuals with similar features need to be removed. The average similarity within a class is proportional to the number of individuals removed within the class, inversely proportional to the number of individuals retained within the class. Taking different intra-class average similarities as a metric, the appropriate numbers of individuals are removed from each class to achieve self-adaptive effect.
Specifically, in the $A$ class, because the hierarchical ranking of the individual $a$ in the non-dominated sorting is the last and away from the Pareto frontier, its existence is not conducive to the combination of new individuals with characteristics that maintain the characteristics of population convergency in the subsequent evolution process. On the contrary, it increases the probability of deviating from the Pareto frontier individuals, resulting in the convergence of the entire population is too slow, the average similarity in this set is relatively small through the adaptive clustering pruning strategy, only the individual $a$ away from the Pareto front surface is removed; in the $D$ class, the individual $b$ has the same situation as the individual $a$, and is removed; in the $C$ class, there are 5 individuals in the class, and the individual $g$ has a relatively large crowding distance, while for the four individuals $c$, $d$, $e$, and $f$, they are in the same non-dominated sorting hierarchy, the features are very similar and the crowding distance is too low, which may result in being eliminated or retained at the same time. If it is eliminated at the same time, the point set in this area will be empty, which is not conducive to the expansion of the entire frontier of Pareto; if it is retained at the same time, it will make the similar individuals in this area too much and the distribution is too dense, which is not conducive to the emergence of new and better individuals with more maintainable population distribution characteristics during the evolution process, which hinders the distribution of the population. The adaptive clustering pruning strategy calculates that the average similarity in this set is relatively large, and the two individuals with poor performance, $c$ and $d$, are removed together (the crowding distance of the individual $g$ is relatively large and belongs to the outstanding individual and will be retained); due to the low intra-class average similarity, the individual collections in class $B$ and class $E$ are all retained by calculation and are not pruned.\\
\indent In summary, the adaptive clustering pruning strategy adaptively removes some non-dominated sorting stratification and individuals with poor crowding distance or away from the ideal frontier for the average similarity of individual sets of similar features. Therefore, the convergency and distribution of the population are well maintained.  
 
\begin{algorithm}[!h]
  \caption{Self-Adaptive Clustering Pruning Strategy}
  \label{code:Adaptive Clustering Pruning Strategy Algorithm}
  \begin{algorithmic}[1]
  \State{Step 1: There are $n$ individuals in the population $P_{new}$ after the end of evolution of a certain generation. The coordinate value of each individual in each dimension is $x_{im}$, which indicates the value of the individual in the $m_{th}$ dimension,  and calculates the value of each target in the problem to be optimized, denoted as $f_{it}$, expressed as the value of the individual on the $t_{th}$ objective function, where $i$ is the sequential number of each individual;}
  \State{Step 2: Calculate the non-dominated sorting stratification of each individual in the population $P_{new}$ according to the superiority and inferior relationship between each target value $f_{it}$, mark the level $c_{i}$ of each individual, and calculate the crowding distance $d_{i}$ of each individual;}
  \State{Step 3:  Based on the coordinate value $x_{im}$ in each dimension of the individual and the value of the non-dominated hierarchical ranking $c_{i}$ and the crowding distance $d_{i}$, the K-means clustering algorithm is used to cluster the individuals in the population to obtain $k$ categories as a set of k individuals with similar features $\{U1, U2...Uk\}$;}
  \State{Step 4: For the individuals in the set $\{U1, U2...Uk\}$ with similar characteristics, calculate the similarity $p_{ij}$ between the two bodies according to the coordinate value $x_{i}$ in each dimension of the individual and the value $f_{it}$ of each target. Where $i, j$ is the sequential number of each individual;}

  \State{ Step 5:  Substitute the similarity $p_{ij}$ between two bodies in the class into the formula \eqref{similarity}, and calculate the average similarity $P_{k}$ in each set};
  \State{Step6. Substitute the average similarity $P_{k}$ in each set into the formula \eqref{number}, and calculate the number $n_k$ of individuals in each class that need to be retained;}

  \State{Step 7: According to the calculated number of individuals $n_{k}$ retained by each class, trim the $(n_{Uk}-n_{k})$ individuals with poor non-dominated hierarchical ranking and crowding distance and the individuals away from the ideal frontier, where $n_{Uk}$ is the number of individuals in the $k_{th}$ set;}

  \State{Step 8:  Reorganize the $n_{k}$ better individuals retained after these pruning into the new population $P_{new}$ and continue the next generation of evolution.}
  \end{algorithmic}
\end{algorithm}
It can be seen from the algorithm flow that in the process of population maintenance, the population distribution is adaptively maintained based on the aggregation characteristics of each set in the population. According to the average similarity of individual sets with similar features, an adaptively pruned a small number of individuals with similar features and non-dominated hierarchical ranking and poor crowding distance, retaining most representative individuals. At the same time, a small number of individuals far from the frontier surface were filtered, which accelerated the rapid convergence of the population to the Pareto frontier during the evolution process. However, the NSGA-II crowding distance solving method is used to complete the selection operation, so that there are a large number of individuals with similar features in the iterative process, in which individuals in each set divided by similar features have a high probability of being homogenous which is not conducive to the maintenance of population distribution; And retaining individual individuals away from the Pareto front, is not conducive to the maintenance of population convergency.

\section{Experiments and discussion}
\subsection{Experimental environment and test functions}
In this experiment, the 26 test functions of SCH, FON, POL, KUR, ZDT series, DTLZ series and UF series of multi-objective optimization problems are used to test the algorithm effect. The program runs under Windows 7 environment of 2.6 GHz CPU and 8G memory.These multi-objective test functions with different characteristics such as continuity, discontinuity, convexity, and concavity are easy to test the optimization effect of the algorithm under different types of problems. Parameter setting: In NSGA-II and OTNSGA-II (in this paper algorithm), population size popsize = 100, running algebra ngen = 250, both algorithms use real number coding, crossover probability $P_m = 0.9$, mutation probability $P_c = 0.1$, To ensure that the comparison experiment was carried out under the same conditions.
\subsection{Performance evaluation}
The two basic evaluation metrics for multi-objective optimization are convergency and distribution. The calculation of the solution set evaluation in multi-objective optimization is an important and complex problem \cite{Ref29}, and the researchers have proposed some effective methods \cite{Ref30-32}.   
For the convergency evaluation, Generational Disdance (GD)\cite{Ref33} can be used to measure the degree of convergence between the final solution set of the algorithm iteration and the Pareto ideal frontier. The formula is as follows:
\begin{equation}
\label{GD}
GD = {1\over n}{\sqrt{\sum\limits_{i=1}^{n}d^2_i}}
\end{equation}
In equation ~\ref{GD}, $n$ is the number of individuals in the solution set, and $d_i$ is the minimum of the Euclidean distance from each individual to each point in the Pareto ideal frontier. The smaller the value of GD, the closer the solution set is to the Pareto ideal frontier. If $GD = 0$, the solution of the algorithm is on the Pareto ideal frontier, which is also the most ideal experimental result.
For the distribution evaluation, the SP \cite{Ref34} proposed by Schott can be used to measure the uniformity of the final solution set distribution of the algorithm iteration. The calculation formula is as follows:
\begin{equation}
\label{SP}
SP = \sqrt{{1\over{n-1}}{\sum\limits_{i=1}^{n}}{(\overline d-d_i)^2}}
\end{equation}
Where $d_i$ represents the distance between adjacent points parallel to the Pareto ideal front, $di =min(|f_{1i} (x) - f_{1j} (x)| + |f_{2i}(x) - f_{2j}(x)|),(i,j = 1,2,...,n)$ ,$\overline d$ is the average of all $d_i$. $SP = 0$ when the final solution set of the algorithm iteration is completely evenly distributed in the target space.   \\
\indent In order to comprehensively evaluate the convergency and distribution of the final solution set of the algorithm iteration, the inverse generation distance (inverted generational distance, referred to as IGD)\cite{Ref35} can be used as a performance evaluation index. IGD is a measure of the distance between the true Pareto frontier and the final solution set of the algorithm iteration. The smaller the value calculated by the index value, the better the convergency and distribution performance of the final solution set of the algorithm iteration, and the closer to the true Pareto frontier and evenly distributed. Let P be the real Pareto optimal solution set, and A is the final solution set of the evolutionary algorithm iteration, then the formula is as follows:
\begin{equation}
\label{IGD}
IGD = {1\over{\left|P\right|}}{\sum\limits_{i=1}^{\left|P\right|}}{Dist_i}
\end{equation}
where,$Dist_i = {{min}_{j-1}^{|A|}}{\sqrt{ {\sum\limits_{m-1}^{M}}{({{f_m(p_i)-f_m(a_j)}\over{f_{m}^{max}-f_{m}^{min}}})^2}}}$ is minimum normalized Euclidean distance;$f_{m}^{max}$and$f_{m}^{min}$ are the maximum and minimum values on the $m_{th}$ target in $P$ ,$m=1,2,...,M$.M is the target number;$p_i\in P,i=1,2,...,|P|;a_j\in A,j=1,2,...,|A|.$

\subsection{Experiment and data analysis}
In the aforementioned experimental environment, using the two different algorithms of NSGA-II and OTNSGA-II (the algorithm of this paper), comparative experiments were performed on the 26 different types of test functions. The experimental results are shown in Table ~\ref{tab:Table1}, Table ~\ref{tab:Table2}, Table ~\ref{tab:Table3}:

\begin{table}[!t]
\centering
\caption{Comparison of experimental results of the GD index of the algorithm convergence is as follows:}
\label{tab:Table1}
\begin{tabular}{|c|l|l|l|}
\hline
Function/Optimal       & Status & NSGAII      & OT-NSGAII  \\ \hline
\multirow{2}{*}{SCH}   & M-best & 0.0016911   & 0.000874   \\ \cline{2-4}
                       & St.dev & $6.447 \times 10^{-4}$  & $4.831 \times 10^{-4}$ \\ \hline
\multirow{2}{*}{FON}   & M-best & 0.001838    & 0.0016056  \\ \cline{2-4}
                       & St.dev & $6.019 \times 10^{-6}$  & $2.482 \times 10^{-5}$ \\ \hline
\multirow{2}{*}{POL}   & M-best & 0.001596    & 0.001452   \\ \cline{2-4}
                       & St.dev & $4.02 \times 10^{-04}$    & $4.53 \times 10^{-07}$   \\ \hline
\multirow{2}{*}{KUR}   & M-best & 0.118343    & 0.01693    \\ \cline{2-4}
                       & St.dev & $6.288 \times 10^{-3}$  & $1.869 \times 10^{-4}$ \\ \hline
\multirow{2}{*}{ZDT1}  & M-best & 0.043195    & 0.001636   \\ \cline{2-4}
                       & St.dev & $1.869 \times 10^{-4}$  & $1.734 \times 10^{-4}$ \\ \hline
\multirow{2}{*}{ZDT2}  & M-best & 0.045597    & 0.00144    \\ \cline{2-4}
                       & St.dev & $4.02 \times 10^{-04}$    & $4.53 times 10^{-07}$   \\ \hline
\multirow{2}{*}{ZDT3}  & M-best & 0.032633    & 0.001595   \\ \cline{2-4}
                       & St.dev & $3.0359 \times 10^{-6}$ & $2.103 \times 10^{-5}$ \\ \hline
\multirow{2}{*}{ZDT4}  & M-best & 0.070377    & 0.012941   \\ \cline{2-4}
                       & St.dev & $4.440 \times 10^{-4}$  & $3.989 \times 10^{-5}$ \\ \hline
\multirow{2}{*}{ZDT6}  & M-best & 0.13382     & 0.001132   \\ \cline{2-4}
                       & St.dev & $1.734 \times 10^{-4}$  & $1.205 \times 10^{-4}$ \\ \hline
\multirow{2}{*}{DTLZ1} & M-best & 1.575658    & 0.620048   \\ \cline{2-4}
                       & St.dev & $2.873 \times 10^{-4}$ & $3.209 \times 10^{-4}$ \\ \hline
\multirow{2}{*}{DTLZ2} & M-best & 0.000442    & 0.000398   \\ \cline{2-4}
                       & St.dev & $1.673 \times 10^{-4}$  & $1.984 \times 10^{-4}$ \\ \hline
\multirow{2}{*}{DTLZ3} & M-best & 3.924024    & 0.411876   \\ \cline{2-4}
                       & St.dev & $2.756 \times 10^{-5}$  & $2.521 \times 10^{-4}$ \\ \hline
\multirow{2}{*}{DTLZ4} & M-best & 0.005644    & 0.005171   \\ \cline{2-4}
                       & St.dev & $2.269 \times 10^{-4}$  & $2.569 \times 10^{-4}$ \\ \hline
\multirow{2}{*}{DTLZ5} & M-best & 0.000232    & 0.000198   \\ \cline{2-4}
                       & St.dev & $1.598 \times 10^{-4}$  & $1.215 \times 10^{-4}$ \\ \hline
\multirow{2}{*}{DTLZ6} & M-best & 0.03898     & 0.0351604  \\ \cline{2-4}
                       & St.dev & $2.986 \times 10^{-4}$  & $2.992 \times 10^{-4}$ \\ \hline
\multirow{2}{*}{DTLZ7} & M-best & 0.005633    & 0.003703   \\ \cline{2-4}
                       & St.dev & $2.521 \times 10^{-4}$  & $2.587 \times 10^{-4}$ \\ \hline
\multirow{2}{*}{UF1}   & M-best & 0.016018    & 0.000878   \\ \cline{2-4}
                       & St.dev & $2.873 \times 10^{-4}$  & $3.209 \times 10^{-4}$ \\ \hline
\multirow{2}{*}{UF2}   & M-best & 0.009346    & 0.000673   \\ \cline{2-4}
                       & St.dev & $1.673 \times 10^{-5}$  & $1.984 \times 10^{-4}$ \\ \hline
\multirow{2}{*}{UF3}   & M-best & 0.015509    & 0.001537   \\ \cline{2-4}
                       & St.dev & $2.756 \times 10^{-4}$ & $2.521 \times 10^{-4}$ \\ \hline
\multirow{2}{*}{UF4}   & M-best & 0.0059      & 0.002664   \\ \cline{2-4}
                       & St.dev & $2.269 \times 10^{-4}$  & $2.569 \times 10^{-4}$ \\ \hline
\multirow{2}{*}{UF5}   & M-best & 0.017519    & 0.002787   \\ \cline{2-4}
                       & St.dev & $1.598 \times 10^{-4}$  & $1.215 \times 10^{-4}$ \\ \hline
\multirow{2}{*}{UF6}   & M-best & 0.423124    & 0.0080702  \\ \cline{2-4}
                       & St.dev & $2.986 \times 10^{-4}$  & $2.992 \times 10^{-4}$ \\ \hline
\multirow{2}{*}{UF7}   & M-best & 0.057867    & 0.01599    \\ \cline{2-4}
                       & St.dev & $2.521 \times 10^{-4}$  & $2.587 \times 10^{-5}$ \\ \hline
\multirow{2}{*}{UF8}   & M-best & 0.415352    & 0.192268   \\ \cline{2-4}
                       & St.dev & $2.218 \times 10^{-4}$  & $2.569 \times 10^{-4}$ \\ \hline
\multirow{2}{*}{UF9}   & M-best & 0.550303    & 0.306235   \\ \cline{2-4}
                       & St.dev & $2.584 \times 10^{-5}$  & $2.794 \times 10^{-5}$ \\ \hline
\multirow{2}{*}{UF10}  & M-best & 1.552098    & 0.962719   \\ \cline{2-4}
                       & St.dev & $4.189 \times 10^{-5}$  & $4.236 \times 10^{-5}$ \\ \hline
\end{tabular}
\end{table}

\begin{table}[!t]
\centering
\caption{Experimental results of the SP index of the algorithm distribution are as follows:}
\label{tab:Table2}
\begin{tabular}{|c|l|l|l|}
\hline
Function/Optimal       & Status & NSGAII      & OT-NSGAII   \\ \hline
\multirow{2}{*}{SCH}   & M-best & 0.86785     & 0.454309    \\ \cline{2-4}
                       & St.dev & $5.697 \times 10^{-4}$  & $3.985 \times 10^{-4}$  \\ \hline
\multirow{2}{*}{FON}   & M-best & 0.424642    & 0.436268    \\ \cline{2-4}
                       & St.dev & $4.569 \times 10^{-5}$  & $5.269 \times 10^{-5}$  \\ \hline
\multirow{2}{*}{POL}   & M-best & 1.002833    & 1.028583    \\ \cline{2-4}
                       & St.dev & $3.91 \times 10^{-04}$    & $6.23 \times 10^{-07}$    \\ \hline
\multirow{2}{*}{KUR}   & M-best & 0.656668    & 0.513558    \\ \cline{2-4}
                       & St.dev & $3.068 \times 10^{-3}$  & $2.351 \times 10^{-4}$  \\ \hline
\multirow{2}{*}{ZDT1}  & M-best & 0.883879    & 0.56739     \\ \cline{2-4}
                       & St.dev & $6.284 \times 10^{-4}$  & $3.254 \times 10^{-4}$  \\ \hline
\multirow{2}{*}{ZDT2}  & M-best & 0.968816    & 0.622617    \\ \cline{2-4}
                       & St.dev & $5.21 \times 10^{-04}$    & $6.26 \times 10^{-07}$    \\ \hline
\multirow{2}{*}{ZDT3}  & M-best & 1.012148    & 0.778596    \\ \cline{2-4}
                       & St.dev & $2.936 \times 10^{-6}$  & $3.065 \times 10^{-6}$  \\ \hline
\multirow{2}{*}{ZDT4}  & M-best & 0.884043    & 0.662301    \\ \cline{2-4}
                       & St.dev & 3.568 $\times 10^{-4}$  & $2.564 \times 10^{-5}$  \\ \hline
\multirow{2}{*}{ZDT6}  & M-best & 1.044825    & 0.65609523  \\ \cline{2-4}
                       & St.dev & $5.264 \times 10^{-5}$  & $5.217 \times 10^{-4}$  \\ \hline
\multirow{2}{*}{DTLZ1} & M-best & 0.952009    & 0.914591    \\ \cline{2-4}
                       & St.dev & $3.0517 \times 10^{-4}$ & $4.021 \times 10^{-5}$  \\ \hline
\multirow{2}{*}{DTLZ2} & M-best & 0.718433    & 0.74509     \\ \cline{2-4}
                       & St.dev & $2.015 \times 10^{-4}$  & $2.364 \times 10^{-4}$  \\ \hline
\multirow{2}{*}{DTLZ3} & M-best & 1.289325    & 0.786008    \\ \cline{2-4}
                       & St.dev & $3.098 \times 10^{-4}$  & $4.261 \times 10^{-4}$  \\ \hline
\multirow{2}{*}{DTLZ4} & M-best & 0.650501    & 0.624583    \\ \cline{2-4}
                       & St.dev & $3.0547 \times 10^{-4}$ & $4.695 \times 10^{-4}$  \\ \hline
\multirow{2}{*}{DTLZ5} & M-best & 0.437614    & 0.3895      \\ \cline{2-4}
                       & St.dev &$ 2.034 \times 10^{-4}$  & $2.0364 \times 10^{-4}$ \\ \hline
\multirow{2}{*}{DTLZ6} & M-best & 0.0761      & 0.0667      \\ \cline{2-4}
                       & St.dev &$ 3.654 \times 10^{-4}$  & $6.548 \times 10^{-4}$  \\ \hline
\multirow{2}{*}{DTLZ7} & M-best & 0.68872     & 0.59462     \\ \cline{2-4}
                       & St.dev & $3.687 \times 10^{-4}$  & $3.215 \times 10^{-4}$  \\ \hline
\multirow{2}{*}{UF1}   & M-best & 1.061791    & 0.459775    \\ \cline{2-4}
                       & St.dev & $2.654 \times 10^{-4}$  & $2.958 \times 10^{-4}$  \\ \hline
\multirow{2}{*}{UF2}   & M-best & 0.520626    & 0.49532     \\ \cline{2-4}
                       & St.dev & $2.354 \times 10^{-4}$  & $2.0359 \times 10^{-4}$ \\ \hline
\multirow{2}{*}{UF3}   & M-best & 1.03532     & 0.488056    \\ \cline{2-4}
                       & St.dev & $3.254 \times 10^{-4}$  & $3.2875 \times 10^{-4}$ \\ \hline
\multirow{2}{*}{UF4}   & M-best & 0.54407     & 0.48887     \\ \cline{2-4}
                       & St.dev & $4.065 \times 10^{-4}$  & $3.687 \times 10^{-4}$  \\ \hline
\multirow{2}{*}{UF5}   & M-best & 1.574736    & 1.34681     \\ \cline{2-4}
                       & St.dev & $2.065 \times 10^{-4}$  & $4.654 \times 10^{-4}$  \\ \hline
\multirow{2}{*}{UF6}   & M-best & 1.36584     & 1.14237     \\ \cline{2-4}
                       & St.dev & $3.057 \times 10^{-4}$ & $3.158 \times 10^{-4}$  \\ \hline
\multirow{2}{*}{UF7}   & M-best & 1.263698    & 1.037796    \\ \cline{2-4}
                       & St.dev & $6.549 \times 10^{-4}$  & $3.542 \times 10^{-4}$  \\ \hline
\multirow{2}{*}{UF8}   & M-best & 0.969654    & 0.941589    \\ \cline{2-4}
                       & St.dev & $3.0024 \times 10^{-4}$ & $2.5697 \times 10^{-4}$ \\ \hline
\multirow{2}{*}{UF9}   & M-best & 0.826774    & 0.998542    \\ \cline{2-4}
                       & St.dev & $3.2601 \times 10^{-5}$ & $2.6541 \times 10^{-5}$ \\ \hline
\multirow{2}{*}{UF10}  & M-best & 0.973165    & 0.941032    \\ \cline{2-4}
                       & St.dev & $6.135 \times 10^{-5}$  & $6.217 \times 10^{-5}$  \\ \hline
\end{tabular}

\end{table}

\begin{table}[!t]
\centering

\caption{The experimental results of the IGD indicators for the comprehensive performance of the algorithm are as follows:}
\label{tab:Table3}
\begin{tabular}{|c|l|l|l|}
\hline
Function/Optimal       & Status & NSGAII      & OT-NSGAII   \\ \hline
\multirow{2}{*}{SCH}   & M-best & 0.007819    & 0.007344    \\ \cline{2-4}
                       & St.dev & $5.0214 \times 10^{-4}$ & $2.3147 \times 10^{-4}$ \\ \hline
\multirow{2}{*}{FON}   & M-best & 0.010146    & 0.009603    \\ \cline{2-4}
                       & St.dev & $9.2647 \times 10^{-6}$ & $1.264 \times 10^{-6}$  \\ \hline
\multirow{2}{*}{POL}   & M-best & 0.012152    & 0.011792    \\ \cline{2-4}
                       & St.dev & $3.12 times 10^{-04}$    & $3.25 \times 10^{-07}$    \\ \hline
\multirow{2}{*}{KUR}   & M-best & 0.845455    & 0.01202     \\ \cline{2-4}
                       & St.dev & $2.657 \times 10^{-3}$  & $1.2367 \times 10^{-4}$ \\ \hline
\multirow{2}{*}{ZDT1}  & M-best & 0.410049    & 0.015877    \\ \cline{2-4}
                       & St.dev & $2.568 \times 10^{-4}$  & $3.245 \times 10^{-4}$  \\ \hline
\multirow{2}{*}{ZDT2}  & M-best & 0.45493     & 0.014074    \\ \cline{2-4}
                       & St.dev & $5.23 \times 10^{-04}$    & $5.16 \times 10^{-04}$    \\ \hline
\multirow{2}{*}{ZDT3}  & M-best & 0.31326     & 0.01193     \\ \cline{2-4}
                       & St.dev & $2.1347 \times 10^{-5}$ & $3.1574 \times 10^{-5}$ \\ \hline
\multirow{2}{*}{ZDT4}  & M-best & 0.680991    & 0.12446     \\ \cline{2-4}
                       & St.dev & $6.2547 \times 10^{-4}$ & $4.0657 \times 10^{-5}$ \\ \hline
\multirow{2}{*}{ZDT6}  & M-best & 0.296516    & 0.003502    \\ \cline{2-4}
                       & St.dev & $6.274 \times 10^{-4}$  & $2.1457 \times 10^{-4}$ \\ \hline
\multirow{2}{*}{DTLZ1} & M-best & 15.487392   & 6.105718    \\ \cline{2-4}
                       & St.dev & $6.687 \times 10^{-4}$  & $5.638 \times 10^{-4}$  \\ \hline
\multirow{2}{*}{DTLZ2} & M-best & 0.003375    & 0.003269    \\ \cline{2-4}
                       & St.dev & $2.657 \times 10^{-4}$  & $2.356 \times 10^{-4}$  \\ \hline
\multirow{2}{*}{DTLZ3} & M-best & 7.04816     & 4.088749    \\ \cline{2-4}
                       & St.dev & $3.254 \times 10^{-4}$  & $2.035 \times 10^{-4}$  \\ \hline
\multirow{2}{*}{DTLZ4} & M-best & 0.043192    & 0.039584    \\ \cline{2-4}
                       & St.dev & $2.135 \times 10^{-4}$  & $2.3157 \times 10^{-4}$ \\ \hline
\multirow{2}{*}{DTLZ5} & M-best & 0.00378356  & 0.001537    \\ \cline{2-4}
                       & St.dev & $2.598 \times 10^{-4}$ & $3.547 \times 10^{-4}$  \\ \hline
\multirow{2}{*}{DTLZ6} & M-best & 0.03898     & 0.0351604   \\ \cline{2-4}
                       & St.dev & $3.257 \times 10^{-4}$  & $2.998 \times 10^{-4}$  \\ \hline
\multirow{2}{*}{DTLZ7} & M-best & 0.038772    & 0.029854    \\ \cline{2-4}
                       & St.dev & $3.664 \times 10^{-4}$  & $2.363 \times 10^{-4}$  \\ \hline
\multirow{2}{*}{UF1}   & M-best & 0.052948    & 0.004042    \\ \cline{2-4}
                       & St.dev & $3.252 \times 10^{-4}$  & $2.539 \times 10^{-4}$  \\ \hline
\multirow{2}{*}{UF2}   & M-best & 0.032155    & 0.003974    \\ \cline{2-4}
                       & St.dev & $2.658 \times 10^{-4}$  & $2.0601 \times 10^{-4}$ \\ \hline
\multirow{2}{*}{UF3}   & M-best & 0.096244    & 0.004227    \\ \cline{2-4}
                       & St.dev & $3.654 \times 10^{-4}$  & $3.0252 \times 10^{-4}$ \\ \hline
\multirow{2}{*}{UF4}   & M-best & 0.047484    & 0.023952    \\ \cline{2-4}
                       & St.dev & $3.256 \times 10^{-4}$  & $3.614 \times 10^{-4}$  \\ \hline
\multirow{2}{*}{UF5}   & M-best & 0.102554    & 0.002791    \\ \cline{2-4}
                       & St.dev & $2.655 \times 10^{-4}$  & $1.325 \times 10^{-5}$  \\ \hline
\multirow{2}{*}{UF6}   & M-best & 0.442601    & 0.119704    \\ \cline{2-4}
                       & St.dev & $3.025 \times 10^{-5}$  & $2.917 \times 10^{-4}$  \\ \hline
\multirow{2}{*}{UF7}   & M-best & 0.189209    & 0.029279    \\ \cline{2-4}
                       & St.dev & $3.181 \times 10^{-4}$  & $2.196 \times 10^{-4}$ \\ \hline
\multirow{2}{*}{UF8}   & M-best & 3.351214    & 1.463192    \\ \cline{2-4}
                       & St.dev & $3.054 \times 10^{-4}$  & $5.264 \times 10^{-4}$ \\ \hline
\multirow{2}{*}{UF9}   & M-best & 3.453525    & 2.118608    \\ \cline{2-4}
                       & St.dev & $3.257 \times 10^{-5}$  & $2.163 \times 10^{-5}$  \\ \hline
\multirow{2}{*}{UF10}  & M-best & 13.2491     & 6.408198    \\ \cline{2-4}
                       & St.dev & $4.065 \times 10^{-5}$  & $3.957 \times 10^{-5}$ \\ \hline
\end{tabular}

\end{table}

\begin{figure}[!h]

  \includegraphics[width=2.5in]{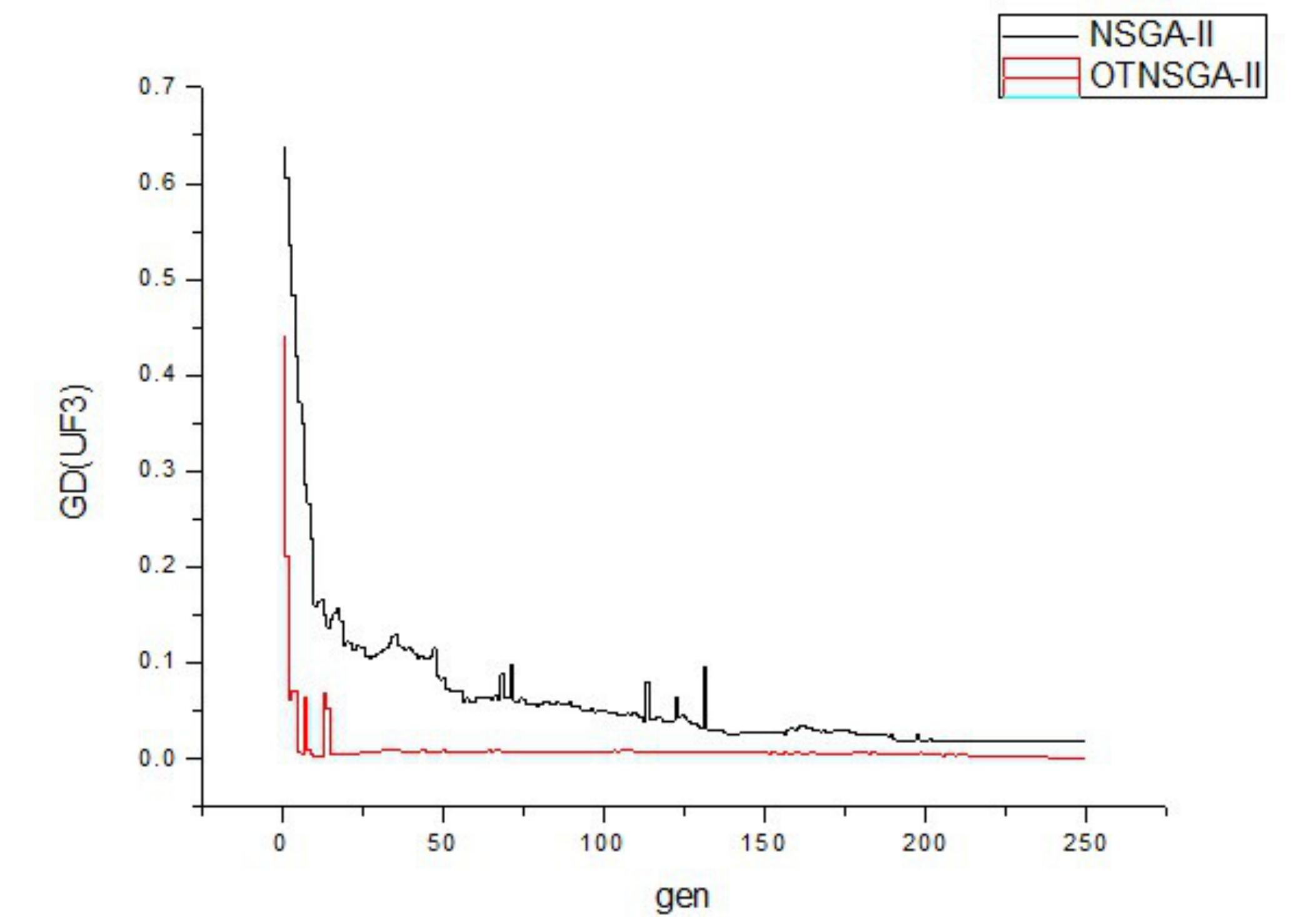}

\caption{Comparison of the {GD} values of the two algorithms.}
\label{fig:GD}       
\end{figure}

\begin{figure}[!h]

  \includegraphics[width=2.5in]{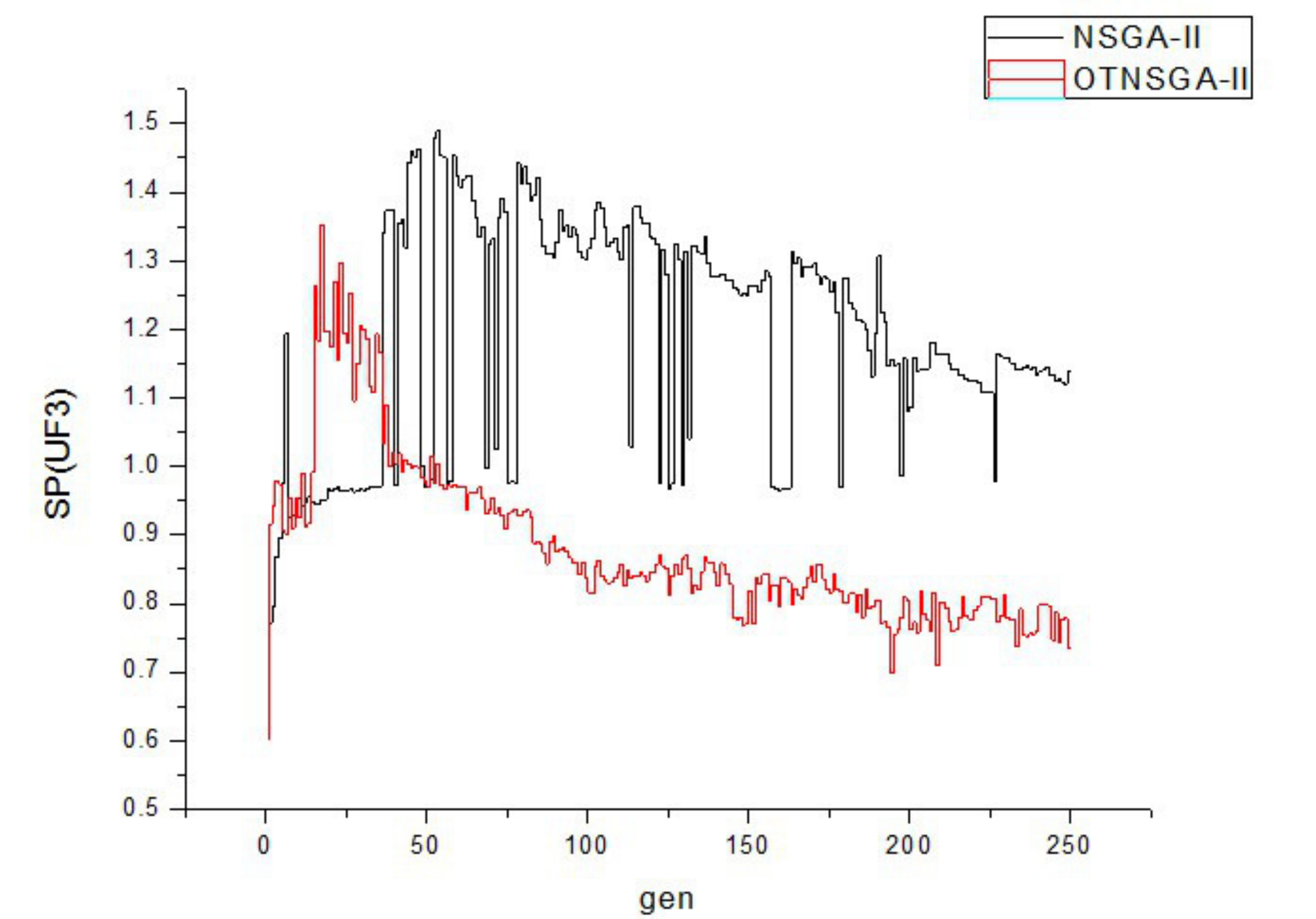}

\caption{Comparison of the {SP} values of the two algorithms.}
\label{fig:SP}    
\end{figure}

\begin{figure}[!h]

  \includegraphics[width=2.5in]{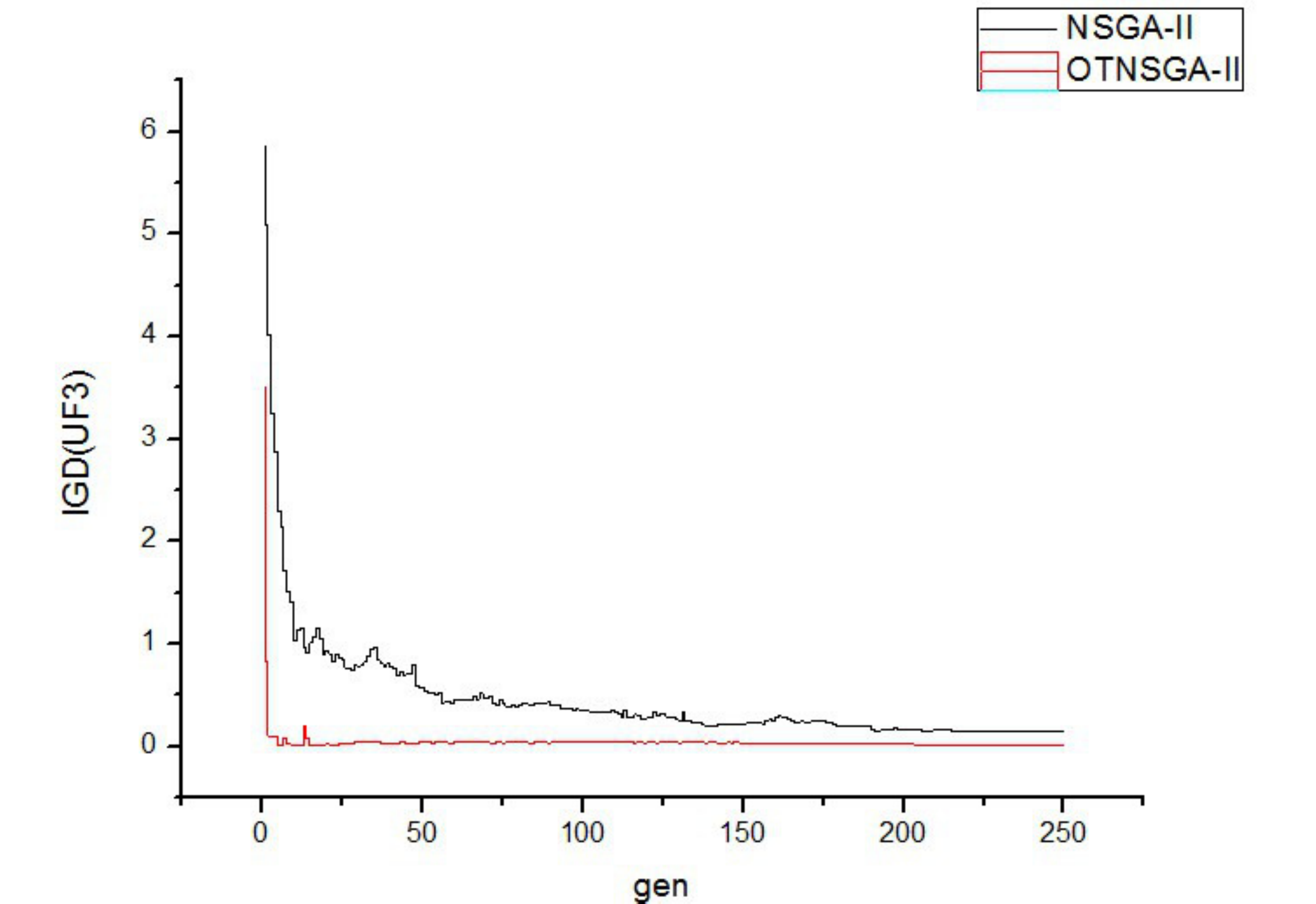}

\caption{Comparison of the {IGD} values of the two algorithms.}
\label{fig:IGD}      
\end{figure}

According to the above experimental results, from the comparison of the convergency indicators in Table~\ref{tab:Table1}, we can find that the convergency of the 26 functions of SCH, FON, POL, KUR, ZDT series, DTLZ series, UF series has been improved, even the 14 functions of SCH, KUR, ZDT1, ZDT2, ZDT3, ZDT6, DTLZ1, DTLZ3, UF1, UF2, UF3, UF5, UF6, UF10 are increased by one order of magnitude or two orders of magnitude. This is because OTNSGA-II uses the method of fault multi-objective orthogonal experiment to initialize the population and select the individual set close to the Pareto ideal front from the evenly distributed individuals as the initial population. This initial advantage of more agile characteristics accelerates the evolutionary trend of subsequent Pareto ideal frontiers; and in the process of evolution, the evolutionary results of each generation are clustered according to the principle of similar individual characteristics within the same class, and adaptively trimming individuals that are farther away from the Pareto ideal frontier in each type of individual set, so that there are more opportunities for the remaining good individuals to combine to produce better individuals, moving closer to Pareto ideals. The direction of the frontier evolved.\\   
\indent  We can find that in the comparison of the distribution indexes in Table~\ref{tab:Table2}, the distribution of 22 functions in the 26 functions of SCH, FON, POL, KUR, ZDT series, DTLZ series and UF series has been improved, Even on the five functions of ZDT3, ZDT6, DTLZ3, UF1, UF3, it is increased by one order of magnitude or two orders of magnitude. It is only slightly worse on the four functions of FON, POL, DTLZ2 and UF9. This is because OTNSGA-II uses the method of fault multi-objective orthogonal experiment to initialize the population, so that the population is evenly distributed in the spatial range closer to the Pareto ideal frontier, maintaining a good initial distribution for the subsequent evolution of the population and laiding the foundation for the population to continue to maintain this good distribution during subsequent evolution; And in the process of evolution, the appropriate number of intra-class non-dominated sorting and individuals with poor crowding distance are trimmed by clustering adaptively, so that most of the individual sets with similar features retain most of the well-represented representative points, which will not make the individual collections of similar features co-exist, thus maintaining a better distribution of the population. However, because the four functions of FON, POL, DTLZ2, and UF9 are characterized by the fact that the Pareto ideal frontier is rather lengthy, In the process of initializing the population by the multi-objective orthogonal experiment of the fault, the method of discretizing the individual by non-dominated sorting and crowding distance is easy to lose some representative points to the maintenance of distribution, and in the adaptive clustering pruning strategy, it is easy to cut off some representative points too much, which will cause a slight loss of distribution, resulting in a slightly worse performance on these four functions.  \\
\indent We can see that in the comparison of the comprehensive performance indicators in Table ~\ref{tab:Table3},the convergence of the 26 functions of SCH, FON, POL, KUR, ZDT series, DTLZ series and UF series has been improved, even in FON, ZDT2, ZDT3. The 11 functions of ZDT6, DTLZ1, UF1, UF2, UF3, UF5, UF7, and UF10 are increased by one order of magnitude or two orders of magnitude. This is because OTNSGA-II uses the method of fault multi-objective orthogonal experiment to initialize the population and select the individual set close to the Pareto ideal front from the uniformly distributed individuals as the initial population, which not only ensures a good initial distribution of the population, but also has the initial convergence of individual individuals in the population close to the Pareto ideal frontier. And adaptively trimming the appropriate number of non-dominated sorts and individuals with poor crowding distance and individuals far from the Pareto ideal frontier in each type of individual set, which ensures that most of the individual collections with similar characteristics retain most of the better representative points, and do not survive the same life, maintaining the distribution of the population; but also filters out a small number of inferior individuals with large deviations, making it easier for the retained good individuals to combine new and better individuals, so that the population accelerates to converge on the Pareto ideal frontier and maintains the convergency of the population.\\
\indent Taking the function UF3 as an example, the detailed changes of the GD, SP, and IGD evaluation index values of the two algorithm populations with algebra are observed, as shown in Figure~\ref{fig:GD},~\ref{fig:SP},~\ref{fig:IGD}.

  In Figure~\ref{fig:GD}, we can see from the comparison of the GD values representing the initial population ($0_{th}$ generation) that the OTNSGA-II design fault multi-objective orthogonal experiment initializes the population so that the generated initial population is close to the ideal frontier. Therefore, the initial convergency of the population is maintained, which is conducive to maintaining this good convergency advantage and accelerating the population convergency in the subsequent evolution process; and the GD value remains stable after 20 generations, which is due to due to the use of adaptive clustering pruning strategy in the evolution process, so that some individuals deviating from the optimal solution set are eliminated, prompting the population to converge rapidly to the Pareto ideal frontier. Before the evolution of NSGA-II ($0_{th}$ generation), the GD value has fallen behind, and it has remained basically stable after 200 years of evolution and is still in a backward state. This is due to the random initialization of the population, which makes the initial abundance of the population not perform well, and because of the retention of individuals away from the frontiers during evolution, the constant participation in the evolution and other individual genes slows down the rapid convergence of the entire population towards the Pareto ideal frontier.

 In Figure ~\ref{fig:SP}, it can be seen from the comparison of the SP values representing the initial population ($0_{th}$ generation) that the initial population of the OTNSGA-II fault multi-objective orthogonal experiment can make the initial population distribution uniform and maintain the initial distribution of the population, which is conducive to the subsequent evolution to continue to maintain this good distribution; And it can be seen from the figure that there is a period of high SP value between the $20_{th}$ and $40_{th}$ generations of the pre-evolution, which is due to the fact that cross-variation between individuals makes the population highly dense. Using the adaptive clustering pruning strategy, the individual collections with similar features will not be the same, leaving most of the representative points in the same set, and removing a few inferior points with similar features and poor performance. The SP value decreased significantly after the $40_{th}$ generation, which is due to the fact that the use of this pruning strategy makes the distribution of the population significantly effective. However, before the evolution of NSGA-II (0th generation), the SP value has lagged behind, and it has gradually fallen behind and become more serious in the evolution process. This is due to the random distribution, which makes the initial distribution poor, and with the continuous evolution of the evolution process, especially after the $50_{th}$ generation, individuals with similar characteristics appear in large numbers and are continuously combined into new individuals with similar characteristics, resulting in high SP values for evaluation distribution, which is not conducive to the maintenance of population distribution.

  In Figure~\ref{fig:IGD}, we can see from the comparison of IGD values representing the initial population ($0_{th}$ generation) that the OTNSGA-II fault multi-objective orthogonal experiment initializes the population to make the initial population distribution uniform and close to the Pareto ideal frontier, and maintains good convergency; And the IGD value remained stable after 20 generations. This is due to the use of the adaptive clustering pruning strategy, which adaptively trims out the non-dominated sorting stratification and the poor crowding distance of individuals with similar characteristics in the same individual set and retains most of the outstanding representative points, so that individuals in similar collections do not simultaneously present or simultaneously pruned, and a small number of individuals away from the front surface are removed, so that the retained excellent individuals produce better individual combinations, thus maintaining the distribution of the population while maintaining the convergency of the population. On the contrary, the IGD value before the evolution of NSGA-II ($0_{th}$ generation) has fallen behind, and it remained stable after the $220_{th}$ generation and remained in a backward state. This is because random initialization makes the initial population distribution uneven and most individuals are far away from the frontier surface, so that they do not have good initial distribution and convergence. In the process of evolution, a large number of individuals are combined with each other to make the population too dense in some spaces, which leads to the simultaneous retention or simultaneous elimination of individuals with similar characteristics, and the existence of some individuals far from the frontier is not conducive to the combination of excellent individuals. So, in the evolution process, it can not maintain good distribution and convergency.

\section{Conclusions}

By analyzing the random population initialization method and the inter-individual crowding distance calculation method in NSGA-II, aiming at the problem that the population distribution and convergency of the algorithm are not properly maintained, this paper proposes an improved algorithm NSNSGA-II of NSGA-II, which uses the method of non-dominated sorting and crowding distance evaluation to set up multi-objective orthogonal experiment to initialize the population. And the iterative results of each generation are clustered by K-means, and some inferior individuals in the class are pruned, and the convergency and distribution of the evolutionary population are maintained. The use of fault multi-objective orthogonal initialization in OTNSGA-II makes the initial individual distribution in the population uniform and close to the Pareto ideal front, maintaining good initial distribution and initial convergence; In the evolution process, the adaptive clustering pruning strategy is used to adaptively prune the appropriate number of individuals with poor crowding distance and non-dominated sorting and individuals far from the Pareto ideal frontier within the same class according to the degree of aggregation within the class, and Prevent individual collections of similar features from being eliminated or retained at the same time, and make the excellent genes easy to retain and develop in the subsequent evolution process, speed up the understanding of the convergence of the set to the Pareto ideal frontier, thus maintaining the distribution and convergency of the population. The comparison experiments between OTNSGA-II and NSGA-II on 26 standard test functions show that the improved algorithm has good distribution performance and convergence performance.

\ifCLASSOPTIONcaptionsoff
  \newpage
\fi

\end{document}